\documentclass[letterpaper]{article} 
\usepackage{aaai25}  
\usepackage{times}  
\usepackage{helvet}  
\usepackage{courier}  
\usepackage[hyphens]{url}  
\usepackage{graphicx} 
\usepackage{natbib}  
\usepackage{caption} 
\usepackage{booktabs}       
\usepackage{multirow, makecell}
\usepackage[table,xcdraw]{xcolor} 
\usepackage{amssymb}
\usepackage{pifont}
\usepackage{amsmath}
\usepackage[ruled,noend,linesnumbered]{algorithm2e}
\usepackage{array}

\title{Look Back for More: Harnessing Historical Sequential Updates for Personalized Federated Adapter Tuning}
\author {
    Danni Peng\textsuperscript{\rm 1},
    Yuan Wang\textsuperscript{\rm 1},
    Huazhu Fu\textsuperscript{\rm 1},
    Jinpeng Jiang\textsuperscript{\rm 2},
    Yong Liu\textsuperscript{\rm 1},
    Rick Siow Mong Goh\textsuperscript{\rm 1},
    Qingsong Wei\textsuperscript{\rm 1}
}
\affiliations {
    \textsuperscript{\rm 1}Institute of High Performance Computing (IHPC),
Agency for Science, Technology and Research (A*STAR), Singapore\\
    \textsuperscript{\rm 2}EVYD Technology\\
    dannip@ihpc.a-star.edu.sg, wang\_yuan@ihpc.a-star.edu.sg, hzfu@ieee.org, jinpeng.jiang@evydtech.com, liuyong@ihpc.a-star.edu.sg, gohsm@ihpc.a-star.edu.sg, wei\_qingsong@ihpc.a-star.edu.sg
}

\usepackage{bibentry}

\begin{document}

\maketitle

\begin{abstract}
Personalized federated learning (PFL) studies effective model personalization to address the data heterogeneity issue among clients in traditional federated learning (FL). Existing PFL approaches mainly generate personalized models by relying solely on the clients' latest updated models while ignoring their previous updates, which may result in suboptimal personalized model learning. To bridge this gap, we propose a novel framework termed \texttt{pFedSeq}, designed for personalizing adapters to fine-tune a foundation model in FL. In \texttt{pFedSeq}, the server maintains and trains a sequential learner, which processes a sequence of past adapter updates from clients and generates calibrations for personalized adapters. To effectively capture the cross-client and cross-step relations hidden in previous updates and generate high-performing personalized adapters, \texttt{pFedSeq} adopts the powerful selective state space model (SSM) as the architecture of sequential learner. Through extensive experiments on four public benchmark datasets, we demonstrate the superiority of \texttt{pFedSeq} over state-of-the-art PFL methods.
\end{abstract}

%

\section{Introduction}

In recent years, federated learning (FL) \cite{mcmahan2017communication} has attracted growing research interest for enabling privacy-preserving collaborative model training. However, due to data heterogeneity among clients (i.e. data from different clients are non-IID or unbalanced), it is difficult to develop a one-fits-all global model that performs well on all clients' local distributions. To address this, personalized federated learning (PFL) \cite{smith2017federated} has emerged. Unlike traditional FL, which develops a single best model for the collective goal, PFL allows each client to have a unique personalized model tailored specifically to the local objective \cite{li2021ditto,fallah2020personalized}. Through the collaborative learning scheme, PFL further enables the personalized models to benefit from knowledge sharing across clients, striking a balance between individualization and generalization for enhanced local performance \cite{kulkarni2020survey,kairouz2021advances}.\

\begin{figure}[t]
\centering
\includegraphics[width=\columnwidth]{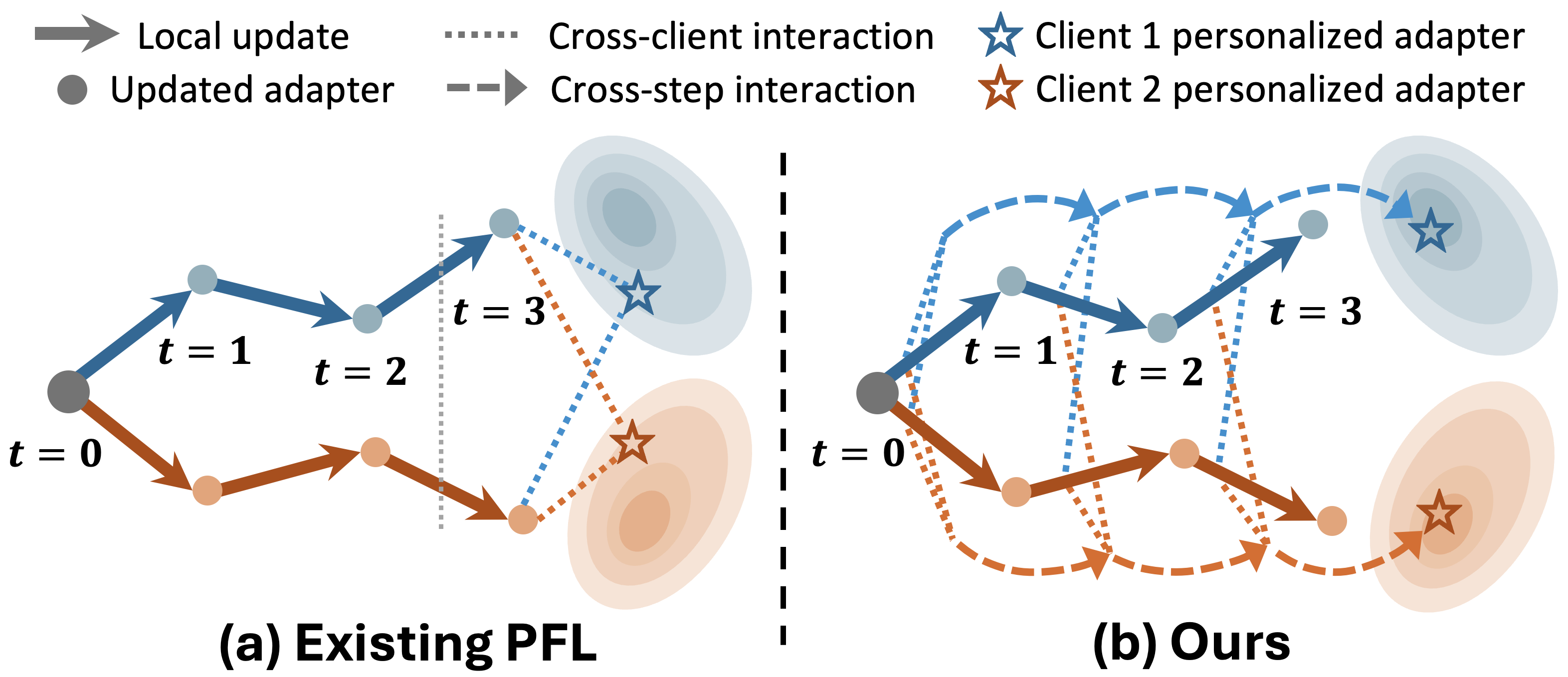} 
\vspace{-5mm}
\caption{Comparison between (a) existing PFL methods and (b) our approach. Instead of leveraging only the latest updates, our approach accounts for past learning trajectories by modeling cross-client and cross-step relations in previous steps, providing a broader view for identifying the consistent trends for learning more robust personalized adapters.}
\label{fig1}
\vspace{-5.5mm}
\end{figure}

Recently, as large foundation models (FMs) demonstrate impressive capabilities across various tasks, there has seen a rise in research synergizing FL and FMs \cite{zhuang2023foundation}. One of the most popular ways of empowering FL with FMs is through parameter-efficient adapter tuning (e.g., LoRA \cite{hu2021lora}), also known as federated adapter tuning \cite{li2024synergizing,woisetschlager2024survey}. This new FL paradigm allows clients to leverage a powerful, pre-trained FM while only fine-tuning and sharing the lightweight adapters with the server, enabling enhanced performance with minimal on-client computation and communication costs. It also opens up opportunities to tailor PFL methods for fine-tuning FM with adapters. For example, leveraging FL for collaborative fine-tuning while personalizing the adapters to better align the representations of pre-trained FM with the local needs of clients \cite{yi2023fedlora,xie2024perada,yang2024dual}.\

However, a common limitation of existing PFL and efforts tailored for FM adapter tuning is that they rely solely on the updates received in the most recent round to develop personalized models, while the potentially valuable information contained in previous updates is either underutilized or completely discarded. This can easily lead to noisy and suboptimal solutions, jeopardizing performance when applied to federated adapter tuning personalization. If we view the development of a personalized adapter for a client as an optimization process, focusing only on the latest updates is akin to performing simple gradient descents. By incorporating information from previous updates—much like using momentum in optimization—the learning process accounts for a broader range of past trajectories and uncovers the consistent trends, leading to more robust and reliable personalized solutions at convergence. Hence, in this work, we are motivated to design a framework that accounts for past learning trajectories to develop enhanced personalized adapters. To illustrate, Figure \ref{fig1} shows the adapter update trajectories of two clients over three rounds of local tuning and communication. As depicted in Figure \ref{fig1}a, existing PFL methods typically rely only on the interactions among the latest updates to produce personalized adapters, which can lead to less robust and undeterministic solutions by only referencing the current states. In contrast, our approach aims to model the interactions among clients at each past update step, as well as the dependencies across various steps, as shown in Figure \ref{fig1}b, which provides a broader perspective for identifying the consistent update patterns, enabling more robust learning and development of more superior personalized adapters. \

Specifically, we propose a novel PFL framework termed \texttt{pFedSeq} for improving personalized federated adapter tuning by exploiting knowledge from clients' past sequential updates. This is achieved through two main processes at the server during each communication round: (1) standard model aggregation using a traditional FL algorithm to obtain a global adapter (e.g., FedAvg \cite{mcmahan2017communication}), and (2) leveraging a \textit{Sequential Learner} to process the sequence of clients' adapter updates collected at the server and output personalized calibrations to adjust the global adapter for clients' individual needs. To enable flexible modeling of the cross-client and cross-step relations in the sequence, we employ a learnable hypernetwork for the sequential learner, which is jointly trained at the server by optimizing the personalized adapters toward clients' local objectives. By using the received adapter updates as a proxy for gradients on local losses, the training of sequential learner can be efficiently performed at server without accessing clients' local data.\

Selecting a suitable architecture for the sequential learner is a non-trivial task. To facilitate effective capture of cross-client and cross-step relations, we propose to employ the selective state space model (SSM) \cite{gu2023mamba} as our sequential learner. Selective SSM is a recurrence-based model introduced recently for efficient sequence modeling. It enjoys both effective performance with time-dependent selectivity and efficient computations with linear scaling. To capitalize on its design features, we form the step-wise inputs by concatenating clients' updates from the same round and processing them in a recurrence mode. With that, the cross-client interactions at different steps can be captured in the time-dependent module parameters, and the sequential processing allows the cross-step dependencies to be captured in the consolidated hidden states.

To summarize, our contributions are as follows:
\begin{itemize}
    \item We propose a novel PFL framework, \texttt{pFedSeq}, for personalizing federated adapter tuning for clients. By leveraging knowledge from the previous updates, \texttt{pFedSeq} generates enhanced personalized adapters which better tailor large FMs' representations to clients' local needs.
    \item We collaboratively train a sequential learner at the server to capture useful cross-client and cross-step relations in the sequential updates, achieving effective personalized adapter generation by adopting Selective SSM as the learner architecture.
    \item We evaluate our \texttt{pFedSeq} rigorously on four large-scale benchmark datasets (i.e., CIFAR-100, Tiny-ImageNet, DomainNet, and Omniglot), and show that our \texttt{pFedSeq} outperforms ten state-of-the-art PFL methods by up to 5.39\%. 
\end{itemize}

\section{Related Work}
Traditional FL algorithms \cite{mcmahan2017communication,li2020federated,karimireddy2020scaffold} following the one-model-fits-all paradigm often suffer from degraded performance on clients' local data in the face of severe data heterogeneity. Recently, personalized FL (PFL) has attracted much attention, which develops customized models to accommodate the diverse needs of clients \cite{mansour2020three}. Generally, efforts in PFL fall under four categories:
\vspace{-0.5mm}

\paragraph{(1) Meta-learning-based methods.} By drawing an analogy between FL and meta-learning \cite{jiang2019improving}, this approach jointly develops a global model from which the personalized models can be effectively fine-tuned with just a few local steps. Per-FedAvg \cite{fallah2020personalized} adopts the spirit of MAML \cite{finn2017model} and jointly learns a global initialization by optimizing for one-step gradient updates. pFedMe \cite{t2020personalized} further allows multiple updates in the inner loop and optimizes a Moreau envelope objective. 
\vspace{-0.5mm}

\paragraph{(2) Personalized-aggregation-based methods.} This line of methods produces personalized models by learning client-specific aggregation weights to combine models from other clients. FedFomo \cite{huang2021personalized} and FedAMP \cite{zhangpersonalized} leverage a rule-based approach to compute weights based on the pair-wise distance between clients' models. APPLE \cite{luo2022adapt} and FedALA \cite{zhang2023fedala} adopt a learning-based approach to optimize the personalized weights directly on clients' local objectives. FedDPA \cite{yang2024dual} was introduced recently, focusing on addressing test-time distribution shifts by combining global and personalized adapters through dynamic weighting for federated adapter tuning.
\vspace{-0.5mm}

\paragraph{(3) Personalized-network-based methods.} This approach develops a personalized model independently for each client while drawing on the general knowledge through various forms of global sharing. FedRep \cite{collins2021exploiting} and FedBN \cite{lifedbn} locally update parts of the network that are sensitive to data distributions (e.g., the classification head or the batch-norm layers), while sharing the rest to leverage the common representations. Ditto \cite{li2021ditto} trains full personalized models locally, using a proximal term to regularize distance from the global model. PerAda \cite{xie2024perada} was recently introduced for federated adapter tuning. It leverages proximal regularization similar to Ditto, while further incorporating knowledge distillation to facilitate information sharing. pFedLoRA \cite{yi2023fedlora} encourages efficient sharing for model-heterogeneous PFL through lightweight adapters.

\vspace{-0.5mm}
\paragraph{(4) Hypernetwork-based methods.} As opposed to solely relying on model aggregation for information sharing, this approach outlines a new way of federation by jointly training a hypernetwork using feedback returned from clients (e.g., the model updates). The hypernetwork is trained to directly generate personalized model parameters based on some inputs. Since it is only trained and maintained at the server, a high-capacity, complex network can be adopted for enhanced diversity of the models generated without concern about the communication costs. pFedHN \cite{shamsian2021personalized} is a pioneering work that leverages this approach to generate personalized models for clients based on the learnable client descriptor vectors. L2C \cite{li2022learning} and pFedLA \cite{ma2022layer} utilize hypernetwork to learn personalized aggregation weights. pFedPG \cite{yang2023efficient} focuses on federated prompt learning and trains a hypernetwork for personalized prompt generation. PeFLL \cite{scott2023pefll} and FedL2P \cite{lee2023fedl2p} condition on clients' local statistics and train a hypernetwork to output personalized models or update strategies.\

Our work, by collaboratively training a sequential learner, falls under the hypernetwork-based methods. Different from the existing works, we propose to generate personalized adapters by leveraging clients' previous updates. In other related fields concerning multi-task/distribution learning similar to FL, past gradient updates are commonly used to enhance task representations \cite{zenke2017continual,flennerhag2018transferring,peng2023clustered} or to derive relations among different parties \cite{yu2020gradient,mansilla2021domain}. In the realm of FL, \cite{ji2019learning} utilizes previous clients' updates to produce global model via an RNN-based aggregator. However, the potential of leveraging previous updates for PFL still remains unexplored. Our work fills this gap by developing a method to extract useful cross-client and cross-step relations from past updates, producing personalized adapters that better suit clients' local specifics.

\section{Method}

\begin{figure*}[t]
\centering
\includegraphics[width=0.95\textwidth]{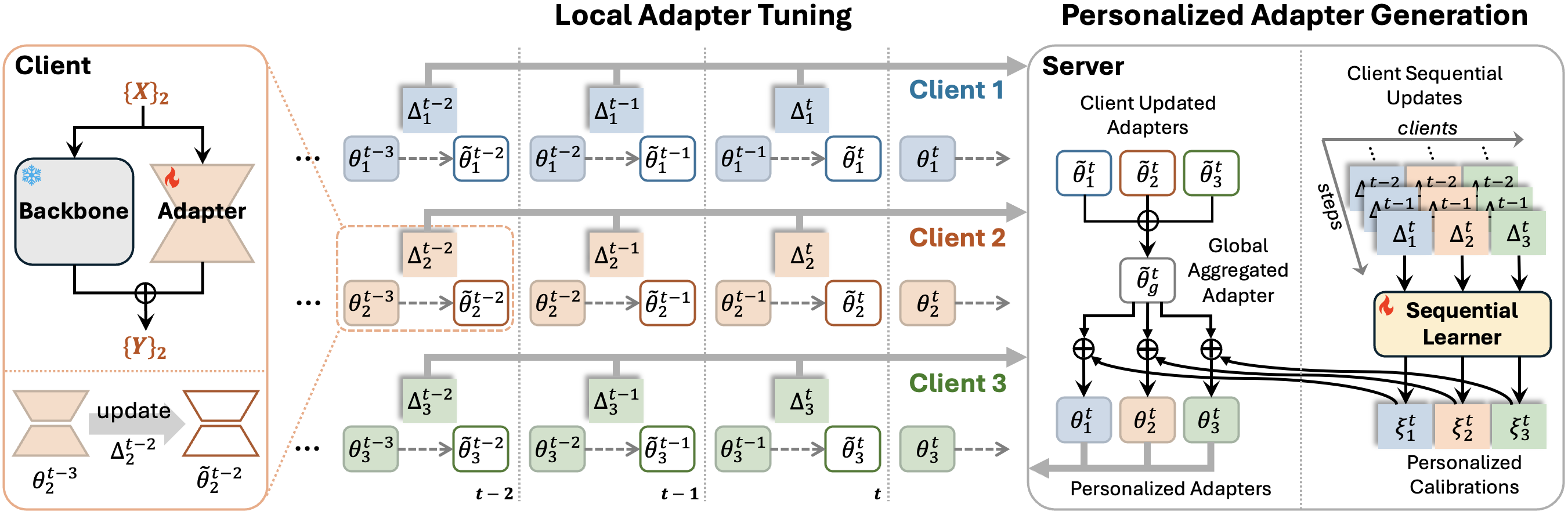} 
\vspace{-1mm}
\caption{An overview of \texttt{pFedSeq} framework. At each communication round $t$, clients perform local adapter tuning and send the adapter updates $\{\Delta_i^t\}_{i=1}^N$ to server. At the server, the updated adapters are aggregated to form a global adapter $\tilde{\theta}_g^t$. Meanwhile, the sequential learner processes the sequence of updates collected at the server and generates personalized calibrations, which are then applied to the global adapter to produce the personalized adapters $\{\theta_i^t\}_{i=1}^N$ and send to the clients.}
\label{fig:overview}
\vspace{-3mm}
\end{figure*}

\subsection{Preliminaries}
\paragraph{Problem Setup.} In a typical FL setup involving $N$ clients and a central server, each client $i \in [N]$ has its own private data $\mathcal{D}_i$. Traditional FL (e.g., FedAvg \cite{mcmahan2017communication}) seeks to learn a global model $\phi_g$ that performs well across all clients: $\min_{\phi_g} \sum_{i=1}^N\mathcal{L}(\mathcal{D}_i;\phi_g)$, where $\mathcal{L}$ is an arbitrary loss function. However, this one-model-fits-all scheme may fail when data heterogeneity is severe among clients. PFL addresses this by relaxing the single-model constraint and learning $N$ personalized models $\{\phi_i\}_{i=1}^N$, each tailored specifically to a client, while also benefiting from the federation by drawing knowledge from global model $\phi_g$:
\begin{equation*}
    \textstyle \min_{\{\phi_i\}_{i=1}^N,\phi_g}\sum_{i=1}^N\mathcal{L}(\mathcal{D}_i;\phi_i,\phi_g).
\label{eqn:pfl_obj}
\end{equation*}

\paragraph{Personalized Federated Adapter Tuning.} Adopting adapter tuning in PFL provides a computation- and communication-efficient solution for clients to harness the power of large FMs. Let $F^*(\cdot)$ denote a fixed, pre-trained foundation model backbone (e.g., ViT \cite{dosovitskiy2020image}), which remains locally at clients. 
A chosen adapter (e.g., LoRA), denoted by $a_{\theta}(\cdot)$ parameterized by $\theta$, as well as the classification head $h_{\omega}(\cdot)$ parameterized by $\omega$, are tuned at the client to adapt the fixed backbone to local distributions. Since only the adapter and the head are updated and shared with the server, without loss of generality, we define the global and personalized models as $\phi_g=(\theta_g, \omega_g)$ and $\phi_i=(\theta_i, \omega_i)$. The individual loss of client $i$ is computed by:
\begin{equation*}
    \textstyle \mathcal{L}(\mathcal{D}_i;\phi_i,\phi_g) := \sum_{(x, y) \in \mathcal{D}_i} l(h_{\omega_i,\omega_g}(F^*(x) + a_{\theta_i,\theta_g}(x)), y).
\end{equation*}

\subsection{\texttt{pFedSeq} Framework}
In this section, we present our \texttt{pFedSeq} framework, designed for personalizing federated adapter tuning. Following \cite{collins2021exploiting,yang2023efficient}, we update the classification head locally without sharing with the server to better preserve the client-specific knowledge, and apply \texttt{pFedSeq} to generate the personalized adapters. \

An overview of \texttt{pFedSeq} is shown in Figure \ref{fig:overview}. At each communication round, \texttt{pFedSeq} performs \textit{local adapter tuning} at the clients and sends the adapter updates to the server. The server then conducts \textit{personalized adapter generation} with the sequential learner by processing the sequence of past updates collected from clients. Meanwhile, the server also performs \textit{sequential learner optimization} using the new updates received from the clients as feedback for training. We now introduce each process in detail.\

\paragraph{Local Adapter Tuning at Client.} 
 Suppose we are at the $t$-th local training round of client $i$. Upon receiving the personalized adapter $\theta_i^{t-1}$ generated by the server in the previous round, we update $\theta_i^{t-1}$ and $\omega_i^{t-1}$ jointly on local data $\mathcal{D}_i$ for several local epochs to obtain $\tilde{\theta}_i^{t}$ and $\omega_i^{t}$. Note that $\omega_i^{t-1}$ is not generated by the server and is restored from the previous local update at client $i$. Also, to be differentiated from the personalized adapter $\theta_i^{t}$ generated by the server, we use $\tilde{\theta}_i^{t}$ to denote the adapter updated locally. We then compute the adapter update $\Delta_i^t := \tilde{\theta}_i^t - \theta_i^{t-1}$ and send it to the server. 

\paragraph{Personalized Adapter Generation at Server.}  
To generate personalized adapters, two processes are carried out at the server: (1) aggregating clients' updated adapters to produce a global adapter, and (2) using the sequential learner to process clients' past updates and output personalized calibrations for tailoring the global adapter to each client.

Specifically, after the $t$-th communication round, server receives adapter updates $\{\Delta_i^t\}_{i=1}^N$ from $N$ clients. To compute the globally aggregated adapter, we need to obtain the clients' updated adapters $\{\tilde{\theta}_i^t\}_{i=1}^N$. To avoid doubling the communication costs, we can compute  $\{\tilde{\theta}_i^t\}_{i=1}^N$ directly using the adapter updates received from the clients and the personalized adapters $\{\theta_i^{t-1}\}_{i=1}^N$ generated at the server in the previous round, i.e., $\tilde{\theta}_i^t = \theta_i^{t-1} + \Delta_i^t, \forall{i\in[N]}$. By using the classic FedAvg \cite{mcmahan2017communication} for aggregation, we obtain the global adapter $\tilde{\theta}_g^t$ at the $t$-th round:
\begin{equation}
    \textstyle \tilde{\theta}_g^t = \sum_{i=1}^N \frac{|\mathcal{D}_i|}{\sum_{i'=1}^N|\mathcal{D}_{i'}|} \tilde{\theta}_i^t.
\label{eqn:global_agg}
\end{equation}

Simultaneously, we utilize a sequential learner, which is a hypernetwork, to generate personalized calibrations from the sequential updates. Specifically, at the $t$-th round, we construct the input to sequential learner by first stacking $\{\Delta_i^{j}\}_{i=1}^N$ across $N$ clients for each $j \in [t]$ collected at the server, forming  $\mathbf{\boldsymbol{\Delta}}^{j}=[\Delta_1^{j},\cdots,\Delta_N^{j}] \in \mathbb{R}^{D\times N}$, where $D$ is the dimensionality of the adapter's parameters. We then concatenate $\{\boldsymbol{\Delta}^{j}\}_{j=1}^{t}$ across $t$ steps, forming the sequence input matrix  $\boldsymbol{\Delta}^{1:t} = [\boldsymbol{\Delta}^{1},\cdots,\boldsymbol{\Delta}^{t}] \in \mathbb{R}^{D \times N \times t}$. The sequential learner, denoted by $\mathrm{SeqLearner}(\cdot;\psi)$ parameterized by $\psi$, outputs the personalized calibrations $\boldsymbol{\xi}^t=[\xi_1^t,\cdots,\xi_N^t]\in\mathbb{R}^{D \times N}$ for the $N$ clients by taking in the sequence input $\boldsymbol{\Delta}^{1:t}$:
\begin{equation}
    \boldsymbol{\xi}^t = \mathrm{SeqLearner}(\boldsymbol{\Delta}^{1:t};\psi).
\label{eqn:inf}
\end{equation}
Note that by treating $D$ as the batch dimension, we employ one $\mathrm{SeqLearner}(\cdot;\psi)$  to capture the cross-client and cross-step relations for all parameters of an adapter. Hence, the size of $\mathrm{SeqLearner}(\cdot;\psi)$ is independent of the adapter size $D$ and only depends on the number of clients $N$ and the sequence length $t$. For better expressivity, we assign one $\mathrm{SeqLearner}(\cdot;\psi)$ to learn the parameters of the adapter attached to each layer of the backbone (e.g., ViT-B/16 contains 12 layers \cite{dosovitskiy2020image}).

Finally, we obtain the personalized adapters $\{\theta_i^t\}_{i=1}^N$ for the $t$-th round by adding $\{\xi_i^t\}_{i=1}^N$ to the global adapter $\tilde{\theta}_g^t$:
\begin{equation}
    \theta_i^t=\tilde{\theta}_g^t+\xi_i^t, \ \ \forall i\in[N].
    \label{eqn:add}
\end{equation}
The personalized adapters $\{\theta_i^t\}_{i=1}^N$ are sent to the clients for the next round of local training.

\begin{figure}[t]
\centering
\includegraphics[width=\columnwidth]{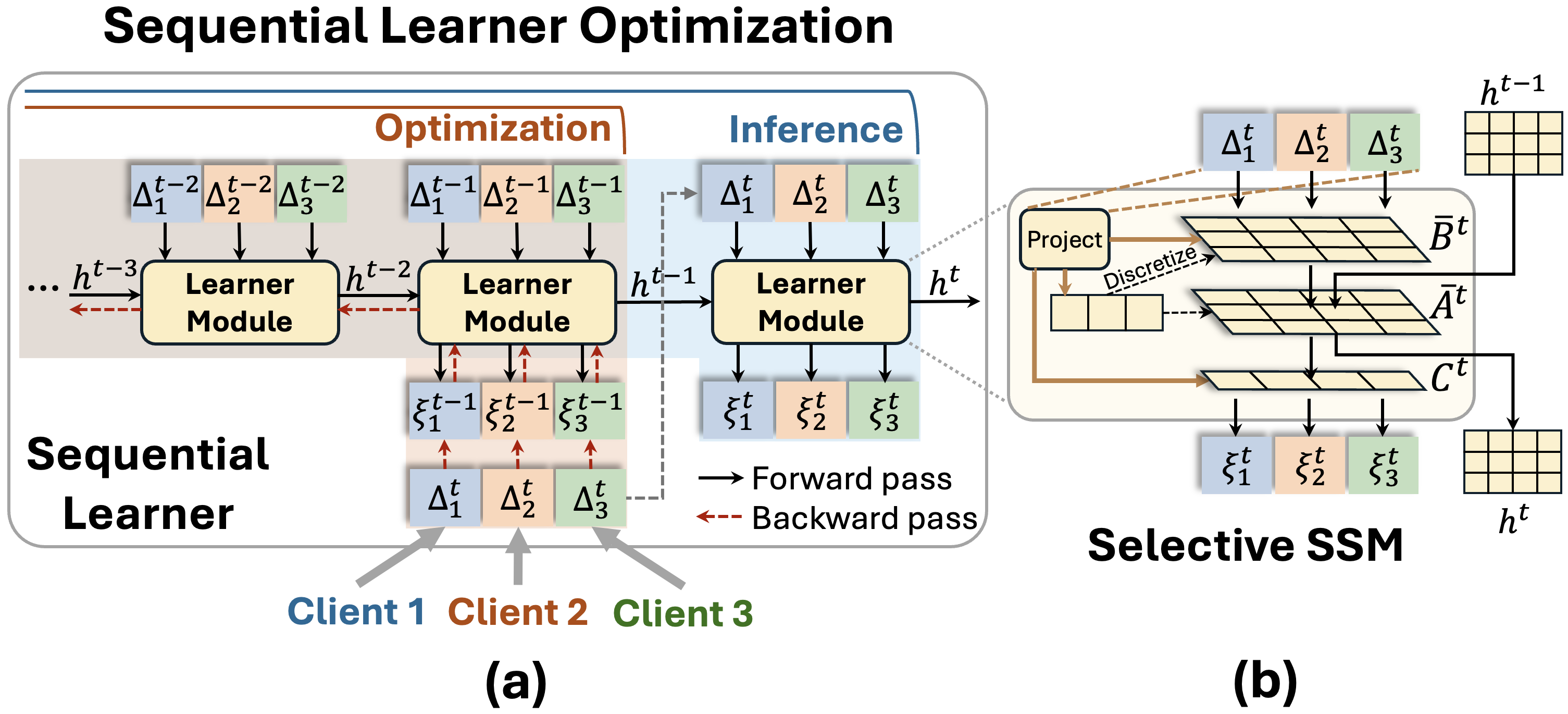} 
\vspace{-5.5mm}
\caption{(a) Optimization and inference processes of sequential learner at the $t$-th round. (b) An instantiation of sequential learner using Selective SSM.}
\label{fig:training}
\vspace{-5mm}
\end{figure}

\paragraph{Sequential Learner Optimization at Server.} 
Since the objective of the sequential learner is to generate effective personalized calibrations that perform well on clients' local data, we optimize the parameters $\psi$ of $\mathrm{SeqLearner}(\cdot;\psi)$ by:
\begin{equation}
\begin{aligned}
     \textstyle \min_{\psi} \textstyle \sum_{i=1}^N  \mathcal{L}(\mathcal{D}_i;\theta_i),
\end{aligned}
\end{equation}
where $\theta_i = \tilde{\theta}_g + \xi_i = \tilde{\theta}_g + (\mathrm{SeqLearner}(\boldsymbol{\Delta};\psi))_{:,i}$. Using chain rule, the gradient update for $\psi$ from each client $i$ is given by $\nabla_{\psi}\mathcal{L}(\mathcal{D}_i;\theta_i)=(\nabla_{\psi}\xi_i)^\top\nabla_{\theta_i}\mathcal{L}(\mathcal{D}_i;\theta_i)$ (note that $\nabla_{\xi_i}\theta_i=\mathbb{I}$). 
Following \cite{shamsian2021personalized,scott2023pefll}, we approximate $\nabla_{\theta_i}\mathcal{L}(\mathcal{D}_i;\theta_i)$ using the adapter update $\Delta_i$ received from client $i$. This is equivalent to replacing a single gradient step on $\theta_i$ with multiple gradient update steps, which has been shown to achieve better and more stable convergence \cite{shamsian2021personalized}.\

Specifically, at the $t$-th round, we receive adapter updates $\{\Delta_i^t\}_{i=1}^N$ from clients. Recall that these are the updates from the previous round's personalized adapters $\{\theta_i^{t-1}\}_{i=1}^N$, i.e., $\Delta_i^t$ is the proxy for gradient evaluated at $\theta_i^{t-1}$, indicating the optimization direction at $\theta_i^{t-1}$. Hence, we use $\{\Delta_i^t\}_{i=1}^N$ as the signal to adjust $\{\theta_i^{t-1}\}_{i=1}^N$, which constitutes $\{\xi_i^{t-1}\}_{i=1}^N$ generated by inputting $\boldsymbol{\Delta}^{1:t-1}$ (i.e., sequence until $t-1$) to $\mathrm{SeqLearner}(\cdot;\psi)$. Formally, we compute the update $(\Delta{\psi})^t$ for $\psi$ at the $t$-th round by:
\begin{equation}
\begin{aligned}
    \textstyle (\Delta{\psi})^t & \textstyle =\sum_{i=1}^N(\nabla_{\psi}\xi_i^{t-1})^\top\Delta_i^t, \\
     & \text{where} \ \ \xi_i^{t-1} = (\mathrm{SeqLearner}(\boldsymbol{\Delta}^{1:t-1};\psi))_{:,i}.
\label{eqn:learner_update}
\end{aligned}
\end{equation}

After updating $\mathrm{SeqLearner}(\cdot;\psi)$, we perform inference to generate $\{\xi_i^t\}_{i=1}^N$ by inputting $\boldsymbol{\Delta}^{1:t}$ (including the latest update $\boldsymbol{\Delta}^{t}$) to $\mathrm{SeqLearner}(\cdot;\psi)$, as described in \eqref{eqn:inf}. The optimization and inference processes of sequential learner at the $t$-th round are illustrated in Figure \ref{fig:training}a.

\paragraph{Remarks.}To avoid infinitely growing sequence length as the number of update rounds increases, we cap the sequence input at a maximum length $L$, i.e., \eqref{eqn:inf} becomes:
\begin{equation}
\boldsymbol{\xi}^t =
\begin{cases} 
      \mathrm{SeqLearner}(\boldsymbol{\Delta}^{1:t};\psi) & \text{for } t \leq L \\
      \mathrm{SeqLearner}(\boldsymbol{\Delta}^{t-L+1:t};\psi) & \text{for } t > L 
\end{cases}
\label{eqn:inf_cap}
\end{equation}
Also, to ensure that $\mathrm{SeqLearner}(\cdot;\psi)$ generates reliable personalized adapters for the next round of local update for stable convergence, we set a warm-up period $W$ to sufficiently train $\mathrm{SeqLearner}(\cdot;\psi)$ before putting it into use. That is, for the first $W$ rounds, the server only updates $\mathrm{SeqLearner}(\cdot;\psi)$ without using it to generate the personalized calibrations, and only the global adapter $\tilde{\theta}_g^t$ is sent to clients during the warm-up period. Algorithm \ref{alg:algo} summarizes the workflow.

\setlength{\textfloatsep}{0pt} 
\begin{algorithm}[tb]
\small
\SetKwInput{KwInput}{Input}                
\SetKwInput{KwOutput}{Output}              
\caption{\texttt{pFedSeq} Framework}
\label{alg:algo}
\KwInput{$N$ clients' local data $\{\mathcal{D}_i\}_{i=1}^N$; Fixed backbone $F^*$; Number of communication rounds $T$; Number of warm-up rounds $W$; Maximum sequence length $L$.}
\KwOutput{Personalized adapters $\{\theta_i^T\}_{i=1}^N$}
\For{$t\in\{1, \cdots, T\}$}
{\textbf{\# Local adapter tuning} \\
\For{client $i \in [N]$ in parallel}
{Update $\theta_i^{t-1},\omega_i^{t-1}$ to $\tilde{\theta}_i^{t},\omega_i^{t}$ by optimizing on $\mathcal{D}_i$.\\
Compute $\Delta_i^t=\tilde{\theta}_i^t - \theta_i^{t-1}$ and send to server.}
\textbf{\# Sequential learner optimization}\\
Perform update on $\psi$ by \eqref{eqn:learner_update}.\\
\textbf{\# Personalized adapter generation}\\
Compute global adapter $\tilde{\theta}_g^t$ by \eqref{eqn:global_agg}.\\
\If{$t \leq W$}
{Set $\theta_i^t=\tilde{\theta}_g^t, \forall i\in[N]$.}
\Else{Compute $\{\theta_i^t\}_{i=1}^N$ by \eqref{eqn:inf_cap} and \eqref{eqn:add}.}
Send personalized adapters $\{\theta_i^t\}_{i=1}^N$ to clients.
}
\end{algorithm}
\setlength{\textfloatsep}{\baselineskip}

\subsection{Sequential Learner using Selective SSM}
In this section, we introduce an instantiation of the sequential learner using Selective SSM as the learner architecture. Selective SSM is recently introduced for efficient sequence modeling \cite{gu2023mamba}. In terms of modeling cross-client and cross-step relations, the selection mechanism of Selective SSM allows cross-client interactions at different steps to be captured in the input-dependent parameters. At the same time, the recursive processing effectively abstracts the cross-step dependencies in the internal hidden states.

Given a sequence input $\boldsymbol{\Delta}^{t-L+1:t} \in \mathbb{R}^{D \times N \times L}$ of length $L$, Selective SSM generates output $\boldsymbol{\xi}^j \in \mathbb{R}^{D \times N}$ at each step $j \in [t-L+1:t]$ by taking in the previous step hidden state $h^{j-1}\in \mathbb{R}^{D \times N \times M}$ and the current step input $\mathbf{\boldsymbol{\Delta}}^{j} \in \mathbb{R}^{D \times N}$, where $M$ is an expanded latent dimension. The step-wise modular operation is formulated as follows:
\begin{equation}
\begin{aligned}
    h^{j} = \bar{\mathbf{A}}^{j}h^{j-1}+\bar{\mathbf{B}}^j\mathbf{\boldsymbol{\Delta}}^{j}, \ \ \ \ \boldsymbol{\xi}^j=\mathbf{C}^j h^{j},
\end{aligned}
\end{equation}
where $\bar{\mathbf{A}}^{j},\bar{\mathbf{B}}^j\in\mathbb{R}^{D\times N \times M}$ are discretized parameters, and $\mathbf{C}^j \in \mathbb{R}^{D\times M}$ is a projection matrix, all obtained by conditioning on the current step input $\boldsymbol{\Delta}^{j}$ (a concatenation of $N$ clients' updates), i.e., $\bar{\mathbf{A}}^{j}=s_{\bar{\mathbf{A}}}(\boldsymbol{\Delta}^{j}),\bar{\mathbf{B}}^{j}=s_{\bar{\mathbf{B}}}(\boldsymbol{\Delta}^{j}),\mathbf{C}^{j}=s_{\mathbf{C}}(\boldsymbol{\Delta}^{j})$ (detailed formulations in \cite{gu2023mamba}). Note that we input a sequence $\boldsymbol{\Delta}^{t-L+1:t}$ into Selective SSM, and optimize or perform inference only on the final-step output $\boldsymbol{\xi}^t$. Figure \ref{fig:training}b illustrates a step module of Selective SSM. \

Following \cite{gu2023mamba}, we incorporate a 1D convolution and a residual connection before and after the Selective SSM, forming a Mamba block. A detailed description of our architecture is included in Appendix A.

\section{Experiments}
\subsection{Experimental Setup}
\paragraph{Datasets and Heterogeneity Scenarios.} We evaluate our \texttt{pFedSeq} on four benchmark datasets covering three different data heterogeneity scenarios. For \textit{label-skew} scenario, we use \textbf{CIFAR-100} \cite{krizhevsky2009learning} which consists of 60,000 images from 100 classes, and \textbf{Tiny-ImageNet} \cite{chrabaszcz2017downsampled} which consists of 110,000 images from 200 classes. For both datasets, we simulate label-skew heterogeneity by distributing data in each class over 10 clients with a Dirichlet distribution $Dir(0.1)$, following \cite{yang2023efficient,zhang2023fedala}. For \textit{feature-skew} scenario, we consider \textbf{DomainNet}, which involves 600,000 images from 345 classes across 6 domains (i.e., Clipart, Infograph, Painting, Quickdraw, Real, and Sketch). Following \cite{lifedbn}, we use the top ten most frequent classes for experiments, simulating feature skew across clients by treating each domain as a client (i.e., $N=6$). Furthermore, we consider a \textit{real-world} heterogeneous scenario where data are collected from actual clients. We adopt \textbf{Omniglot}, which contains images of 1,623 characters from 50 alphabets, handwritten by 20 different individuals. We treat each individual as a client (i.e., $N=20$) and predict the alphabet to which a character belongs.

\vspace{-1mm}
\paragraph{Baselines.} We compare the performance of \texttt{pFedSeq} against 12 baselines, including 2 traditional methods and 10 state-of-the-art PFL methods. For traditional baselines, we consider \textbf{Local} and classic \textbf{FedAvg}, where the former performs local adapter tuning only without sharing information with other clients, while the latter aggregates a global adapter and a global head to share with all clients. For PFL baselines, we include meta-learning-based methods \textbf{Per-FedAvg} \cite{fallah2020personalized} and \textbf{pFedMe} \cite{t2020personalized}; personalized-aggregation-based methods \textbf{APPLE} \cite{luo2022adapt} and \textbf{FedALA} \cite{zhang2023fedala}; personalized-network-based methods \textbf{FedRep} \cite{collins2021exploiting}, \textbf{Ditto} \cite{li2021ditto}, and \textbf{PerAda} \cite{xie2024perada}; and hypernetwork-based methods \textbf{pFedHN} \cite{shamsian2021personalized}, \textbf{pFedLA} \cite{ma2022layer}, and \textbf{PeFLL} \cite{scott2023pefll}. All methods are evaluated on clients' local test sets, and the final result is computed by averaging over all clients.

\vspace{-1mm}
\paragraph{Implementation Details.} For fair comparisons, we adapt all compared methods to the same federated adapter tuning setup, where we adopt ViT-B/16 \cite{dosovitskiy2020image} pre-trained on ImageNet21k \cite{deng2009imagenet} as the fixed backbone $F^*$ and fine-tune LoRA adapter \cite{hu2021lora}. We implement all methods by tuning and sharing only the LoRA and the classification head using the respective PFL algorithms. For all datasets, the number of communication rounds $T$ is set to 80. At each round, the clients perform adapter tuning for 1 local epoch using SGD optimizer with a batch size of 32. The local learning rate is set to 0.05 for Omniglot and 0.005 for other datasets. For our \texttt{pFedSeq}, we adopt a 2-layer Mamba as our sequential learner and set the expanded state dimension $M$ to 16. We train our sequential learner using Adam optimizer with learning rate 0.001, similarly for other hypernetwork-based methods. For all datasets, we set the number of warm-up rounds $W$ to 10 and tune the maximum sequence length $L$ in \{5, 10, 15, 20, 25, 30\}. All experiments are conducted on NVIDIA A100 GPUs with 40GB memory. We repeat each experiment with 3 seeds and report the mean and standard deviation. More implementation details can be found in Appendix B.

\vspace{-1mm}
\subsection{Baseline Comparison}

\begin{table}[t]
\setlength{\tabcolsep}{2pt}
\small
\centering
\begin{tabular}{l | c c c c}
\toprule
\multirow{2}{*}{\small\textbf{Method}} & \multicolumn{2}{c}{\scriptsize\textit{Label-Skew}} & \scriptsize\textit{Feature-Skew} & \scriptsize\textit{Real-World} \\
\cmidrule(lr){2-3} \cmidrule(lr){4-4} \cmidrule(lr){5-5}
&  \small CIFAR-100 & \small Tiny-ImageNet & \small DomainNet & \small Omniglot \\
\midrule
Local & 92.94\tiny$\pm$0.02 & 92.52\tiny$\pm$0.11 & 78.67\tiny$\pm$0.15 & 38.36\tiny$\pm$0.24 \\
FedAvg & 87.36\tiny$\pm$0.12 & 87.98\tiny$\pm$0.19 & 78.19\tiny$\pm$0.19 & 37.87\tiny$\pm$0.84\\
\midrule
\rowcolor{gray!20} \multicolumn{5}{l}{\scriptsize\textit{Meta-Learning-Based}} \\
Per-FedAvg & 93.68\tiny$\pm$0.08 & 93.28\tiny$\pm$0.03 & 80.62\tiny$\pm$0.67 & 40.73\tiny$\pm$0.40\\
pFedMe & 93.35\tiny$\pm$0.10 & 93.62\tiny$\pm$0.04 & 81.71\tiny$\pm$0.28 & 41.90\tiny$\pm$0.44\\
\midrule
\rowcolor{gray!20} \multicolumn{5}{l}{\scriptsize\textit{Personalized-Aggregation-Based}} \\
APPLE & 93.38\tiny$\pm$0.08 & 93.23\tiny$\pm$0.01 & 80.34\tiny$\pm$0.81 & 40.35\tiny$\pm$0.72\\
FedALA & 93.73\tiny$\pm$0.02 & 93.30\tiny$\pm$0.02 & 81.17\tiny$\pm$0.76 & 39.86\tiny$\pm$0.46\\
\midrule
\rowcolor{gray!20} \multicolumn{5}{l}{\scriptsize\textit{Personalized-Network-Based}} \\
FedRep & 94.06\tiny$\pm$0.11 & 93.01\tiny$\pm$0.18 & 81.31\tiny$\pm$0.48 & 38.04\tiny$\pm$0.10\\
Ditto & 93.87\tiny$\pm$0.05 & 93.46\tiny$\pm$0.02 & 80.74\tiny$\pm$0.90 & 41.18\tiny$\pm$0.46\\
PerAda & 93.91\tiny$\pm$0.07 & 93.43\tiny$\pm$0.02 & 80.93\tiny$\pm$0.85 & 41.72\tiny$\pm$0.53\\
\midrule
\rowcolor{gray!20} \multicolumn{5}{l}{\scriptsize\textit{Hypernetwork-Based}} \\
pFedHN & \underline{94.46}\tiny$\pm$0.37 & 93.51\tiny$\pm$0.18 & 81.98\tiny$\pm$0.40 & 42.48\tiny$\pm$0.19\\
pFedLA &  93.42\tiny$\pm$0.21 & 93.37\tiny$\pm$0.11 & 80.26\tiny$\pm$0.20 & 41.27\tiny$\pm$0.55\\
PeFLL & 94.41\tiny$\pm$0.05 & \underline{93.70}\tiny$\pm$0.26 & \underline{82.42}\tiny$\pm$0.17 & \underline{42.98}\tiny$\pm$0.18\\
\midrule
pFedSeq & \textbf{95.30}\tiny$\pm$0.09 & \textbf{94.30}\tiny$\pm$0.08 & \textbf{84.63}\tiny$\pm$0.24 & \textbf{45.25}\tiny$\pm$0.16\\
\bottomrule
\end{tabular}
\vspace{-2mm}
\caption{Performance comparison on four datasets. For fair comparisons, all methods are adapted to the same federated adapter tuning setup (i.e., fine-tuning fixed ViT with LoRA). }
\label{tbl:overall}
\vspace{-2mm}
\end{table}

Table \ref{tbl:overall} compares the performance of \texttt{pFedSeq} against baselines applied to the same federated adapter tuning setup. On all the four datasets, our \texttt{pFedSeq} achieves the highest performance, with significant margins of \{0.84\%, 0.6\%, 2.21\%, 2.27\%\} over the second-best performer (i.e., pFedHN or PeFLL). Notably, we observe greater gains for DomainNet and Omniglot characterized by distinct domain discrepancies across clients. This demonstrates the efficacy of our method in personalizing adapters to better adjust the backbone representations for different styles or feature distributions. Throughout Table \ref{tbl:overall}, we observe that all PFL methods outperform Local and FedAvg, indicating their effectiveness in balancing global and local knowledge. As compared to meta-learning and personalized-network methods, which drive the personalized adapters too closely to the global model (through common initialization or regularization), our \texttt{pFedSeq} directly produces diverse personalized calibrations that better adjust the global adapter to clients' local specifics, leading to better results. Unlike aggregation-based methods, which linearly combine client-specific adapters, \texttt{pFedSeq} leverages the non-linearity of a hypernetwork to model the complex client relations and achieves better knowledge transfer. A closer look at Table \ref{tbl:overall} reveals that hypernetwork-based methods, such as pFedHN, PeFLL, and our \texttt{pFedSeq}, outshine other types of methods, demonstrating the advantages of using hypernetworks to directly produce small-sized adapters. Our \texttt{pFedSeq}, by using the sequential learner to capture the cross-client and cross-step relations, further outperforms the existing hypernetwork-based methods. More discussions on baseline comparisons are included in Appendix C. \

\begin{figure}[t]
\centering
\includegraphics[width=\columnwidth]{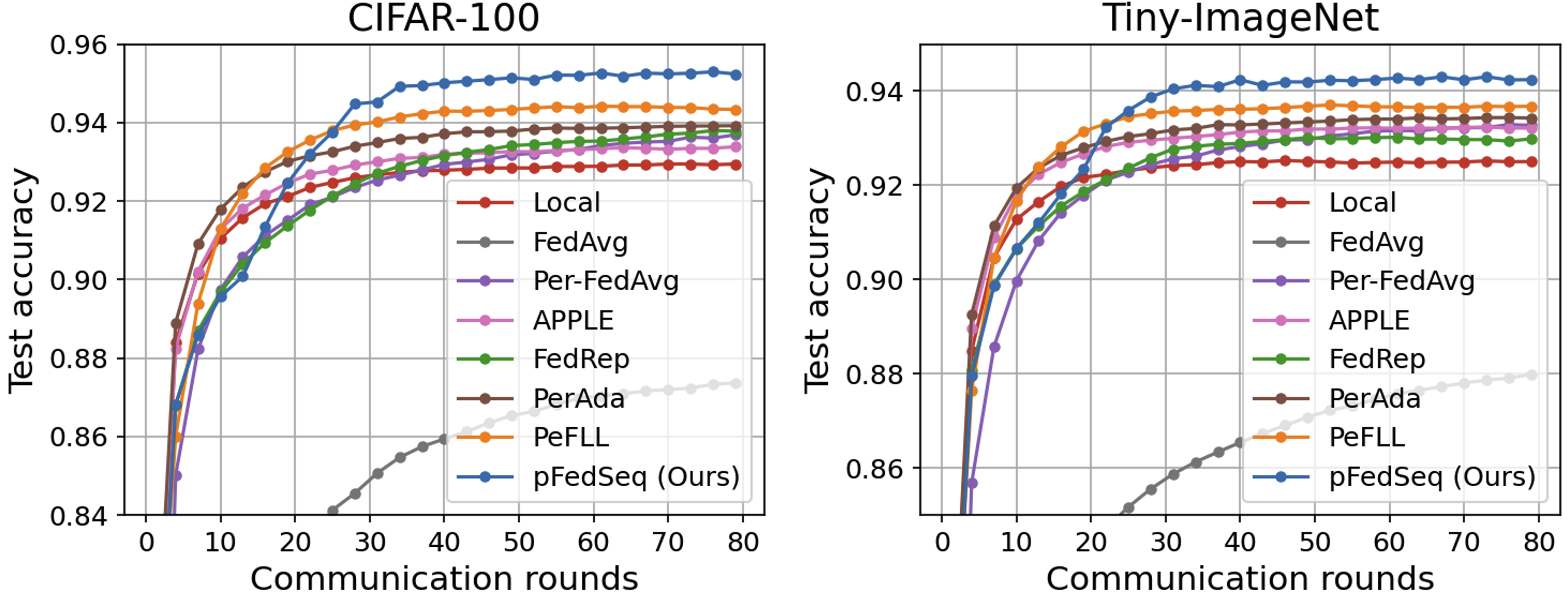} 
\vspace{-6mm}
\caption{Learning curves of \texttt{pFedSeq} and compared baselines for CIFAR-100 and Tiny-ImageNet.}
\label{fig:overall_perf_2}
\vspace{-3mm}
\end{figure}

In Figure \ref{fig:overall_perf_2}, we show the learning curves of \texttt{pFedSeq} and compared baselines over 80 communication rounds for CIFAR-100 and Tiny-ImageNet (the plots for DomainNet and Omniglot are included in Appendix C). For clearer visualization, we only present the curves of representative baselines under each PFL category. From the plots, we can see that our \texttt{pFedSeq} (blue line) begins to outperform all the baselines at around the 25-th round. Though our \texttt{pFedSeq} shows similar performance to FedRep (green line) during the warm-up phase (i.e., the first 10 rounds), where only the global adapter is sent out for clients' local evaluation, we observe rapid improvement once \texttt{pFedSeq} begins to use the personalized adapters generated by the stably trained sequential learner, quickly surpassing all baselines in 15 rounds. The accelerated learning demonstrates the advantages of leveraging previous update steps for deriving more robust and superior personalized adapters, which enables a positive feedback loop between enhanced local updates and better generated personalized adapters, facilitating faster learning and improved performance at convergence.

\vspace{-1mm}
\subsection{Analysis of \texttt{pFedSeq}}
In this section, we first conduct experiments to verify the effectiveness of the various design components of \texttt{pFedSeq}. Then, we provide detailed analysis of the impact of the maximum sequence length $L$ on the performance of \texttt{pFedSeq}.

\vspace{-0.5mm}
\paragraph{Effectiveness of Key Components.} We examine how the three key components of \texttt{pFedSeq} (i.e., global aggregation, cross-step modeling, cross-client modeling) contribute to its overall performance by introducing three variants: variant A removes global aggregation from \texttt{pFedSeq} and uses the sequential learner to generate personalized adapters directly instead of personalized calibrations; variant B eliminates cross-step modeling by using only the latest updates from all clients to generate personalized calibrations, equivalent to setting $L=1$;  variant C eliminates cross-client modeling by generating the personalized calibration for a client using only that client's sequential updates.\

As shown in Table \ref{tbl:key_component}, we see that removing global aggregation results in the largest performance drops: 0.89\% for CIFAR-100 and 1.96\% for DomainNet, signifying the importance of directly leveraging the knowledge sharing through the global adapter. Also, we note that even without global aggregation, variant A performs comparably to the strongest baselines (e.g., pFedHN and PeFLL), which shows the effectiveness of the cross-step and cross-client modeling for directly generating the personalized adapters. Next, we see that removing either cross-step or cross-client modeling leads to significant performance drops on both datasets: 0.64\% and 0.9\% for CIFAR-100, and 1.53\% and 0.78\% for DomainNet, indicating the importance of taking into account both and exploiting their coupled effects in our design. In addition, we observe that both variants B and C surpass the strongest baselines on the two datasets, which shows that even when considered individually, either cross-step or cross-client modeling is effective for generating enhanced personalized adapters. 

\setlength{\tabcolsep}{2.5pt}
\begin{table}[t]
\small
\centering
\begin{tabular}{c | c c c | c c }
\toprule
\multirowcell{2}{\textbf{Variant}} & \multirowcell{2}{\scriptsize global \\ \scriptsize aggregation} & \multirowcell{2}{\scriptsize cross-step\\ \scriptsize modeling} & \multirowcell{2}{\scriptsize cross-client \\ \scriptsize modeling}  & \multirowcell{2}{CIFAR-100} & \multirowcell{2}{DomainNet}\\
& & & \\
\midrule
A & \ding{55} & \ding{51} & \ding{51} & 94.41\tiny$\pm$0.11 & 82.67\tiny$\pm$0.26 \\
B & \ding{51} & \ding{55} & \ding{51}& 94.66\tiny$\pm$0.08 & 83.10\tiny$\pm$0.36 \\
C & \ding{51} & \ding{51} & \ding{55} & 94.40\tiny$\pm$0.12 & 83.85\tiny$\pm$0.23 \\
\midrule
pFedSeq & \ding{51} & \ding{51} & \ding{51} & \textbf{95.30}\tiny$\pm$0.09 & \textbf{84.63}\tiny$\pm$0.24 \\
\bottomrule
\end{tabular}
\vspace{-2.5mm}
\caption{Performance of variants with individual key components removed from \texttt{pFedSeq}.}
\label{tbl:key_component}
\vspace{-3mm}
\end{table}

\setlength{\tabcolsep}{6pt}
\begin{table}[t]
\small
\centering
\begin{tabular}{c | c | c c }
\toprule
\multirowcell{2}{\textbf{Variant}} & \multirowcell{2}{ \textit{Learner} \\ \textit{Architecture}}  & \multirowcell{2}{CIFAR-100} & \multirowcell{2}{DomainNet} \\
&
\\
\midrule
D & MLP & 94.32\tiny$\pm$0.17 & 81.25\tiny$\pm$0.56 \\
E& LSTM & 94.69\tiny$\pm$0.23 & 82.77\tiny$\pm$0.36 \\
\midrule
pFedSeq & Selective SSM & \textbf{95.30}\tiny$\pm$0.09 & \textbf{84.63}\tiny$\pm$0.24 \\
\bottomrule
\end{tabular}
\vspace{-2mm}
\caption{Performance of variants using different architectures for the sequential learner.}
\label{tbl:learner_archi}
\vspace{-2.5mm}
\end{table}

\vspace{-1.5mm}
\paragraph{Effectiveness of Learner Architecture.} To verify our choice of using Selective SSM as the sequential learner, we further introduce two variants using different architectures. First, variant D employs an MLP-based network similar to \cite{shamsian2021personalized} for the sequential learner, where the sequence of inputs are concatenated along a single dimension (i.e., $\boldsymbol{\Delta}^{t-L+1:t} \in \mathbb{R}^{D\times(N\times L)}$) and passed into the MLP network. Note that this architecture can only process fixed-length sequences. As shown in Table \ref{tbl:learner_archi}, using an MLP learner leads to performance drops of 0.98\% for CIFAR-100 and 3.38\% for DomainNet. The decline is mainly attributed to the less effective structure of MLP for modeling sequence inputs, leading to instability of training (as can be observed in Figure \ref{fig:learner_archi_curves} in Appendix C). As for the second architecture, variant E utilizes the classical LSTM \cite{hochreiter1997long} for the learner. Similar to Selective SSM, LSTM is capable of modeling variable-length sequences. However, it reuses the same cell for all the steps without selectivity, making it less capable of discerning valuable context information from different steps. Table \ref{tbl:learner_archi} shows that using an LSTM learner leads to performance declines of 0.61\% on CIFAR-100 and 1.86\% on DomainNet. The less satisfactory performance of variants D and E on two datasets demonstrates the superiority of using Selective SSM to model the cross-client and cross-step dependencies for more effective personalization.

\begin{figure}[t]
\centering
\includegraphics[width=0.9\columnwidth]{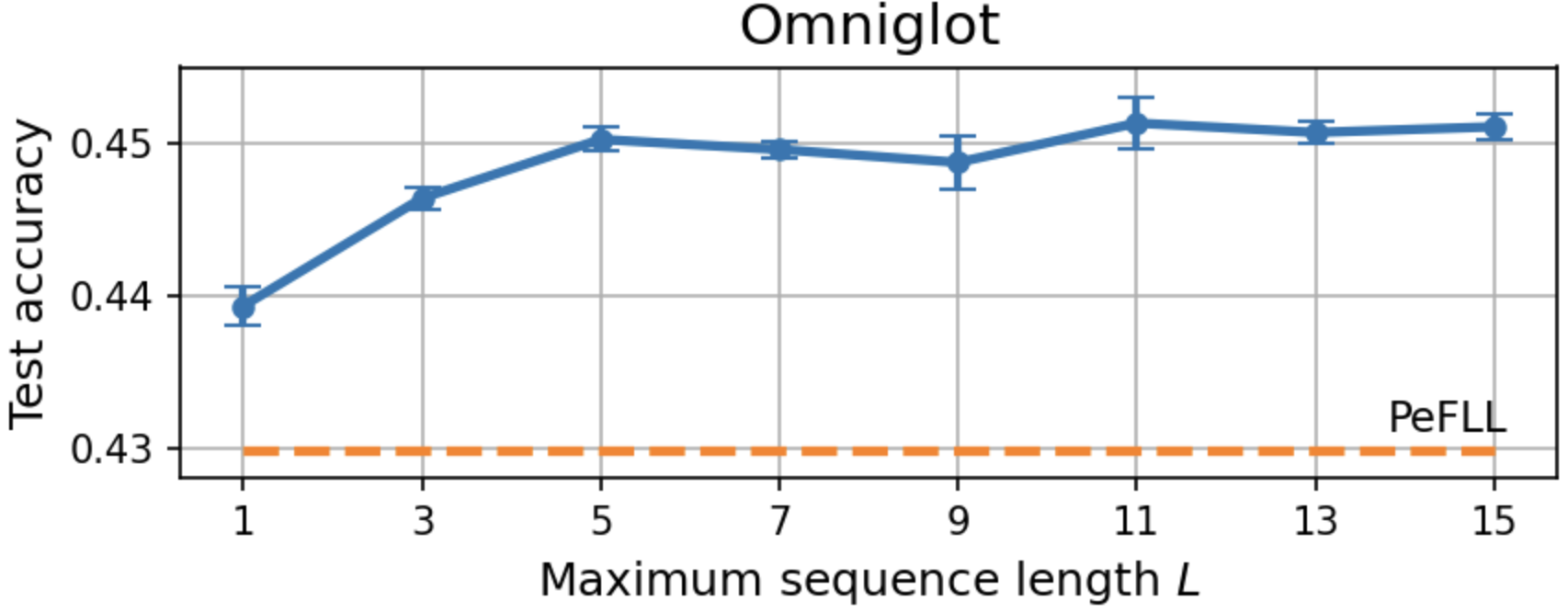} 
\vspace{-2mm}
\caption{Performance of \texttt{pFedSeq} by varying  the maximum sequence length $L$ on Omniglot.}
\vspace{-3mm}
\label{fig:seq_plot}
\end{figure}

\vspace{-1mm}
\paragraph{Impact of Maximum Sequence Length $L$.} We investigate the impact of $L$ by varying it in a more fine-grained range \{1, 3, 5, 7, 9, 11, 13, 15\} on Omniglot dataset. The results are plotted in Figure \ref{fig:seq_plot}, where the dotted line indicates the performance of the strongest baseline PeFLL as a reference. From the plot, we can clearly observe an increasing trend in the performance of \texttt{pFedSeq} as the maximum sequence length $L$ increases. As compared to modeling only on the latest updates (i.e., $L=1$), modeling on a longer sequence of previous updates (i.e., $L=15$) results in 1.17\% increase in the performance of \texttt{pFedSeq}. This further confirms the effectiveness of our approach in modeling previous sequential updates. Moreover,  the performance appears to plateau at a certain level, with further increases in 
$L$ yielding minimal improvements. This may be because earlier updates that are too distant from the present offer less relevant information, making them less useful for generating personalized adapters for the next update round. To achieve the best trade-off between computational cost and performance, we tune $L$ to find the elbow point where the performance starts to plateau. Overall, our \texttt{pFedSeq} outperforms PeFLL even with $L=1$, signifying the effectiveness of our choice of architecture in generating personalized calibrations, and our global aggregation process for explicitly leveraging the global knowledge. A plot of the learning curves for different values of $L$ is included in Appendix C.

\vspace{-2mm}
\section{Conclusion}
In this paper, we propose a novel \texttt{pFedSeq} framework for personalizing federated adapter tuning by exploiting knowledge from clients' previous updates. Our \texttt{pFedSeq} introduces a sequential learner jointly trained across all clients at the server to capture the cross-client and cross-step relations from the sequential updates, and output effective personalized adapters. For the learner architecture, we employ the powerful Selective SSM to leverage its sequence modeling capabilities. Extensive experiments on four benchmark datasets demonstrate the superiority of \texttt{pFedSeq} over ten state-of-the-art PFL methods and verify the effectiveness of its various components through rigorous studies.

\section{Acknowledgments}
This research is supported by the RIE2025 Industry Alignment Fund – Industry Collaboration Project (IAF-ICP) (Award No: I2301E0020), administered by A*STAR.

\bibliography{aaai25}

\clearpage
\appendix

\section{A. Model Architecture of Sequential Learner}
\subsection{Full Architecture}
\label{appx:full_archi}

\begin{figure}[ht]
\centering
\includegraphics[width=\columnwidth]{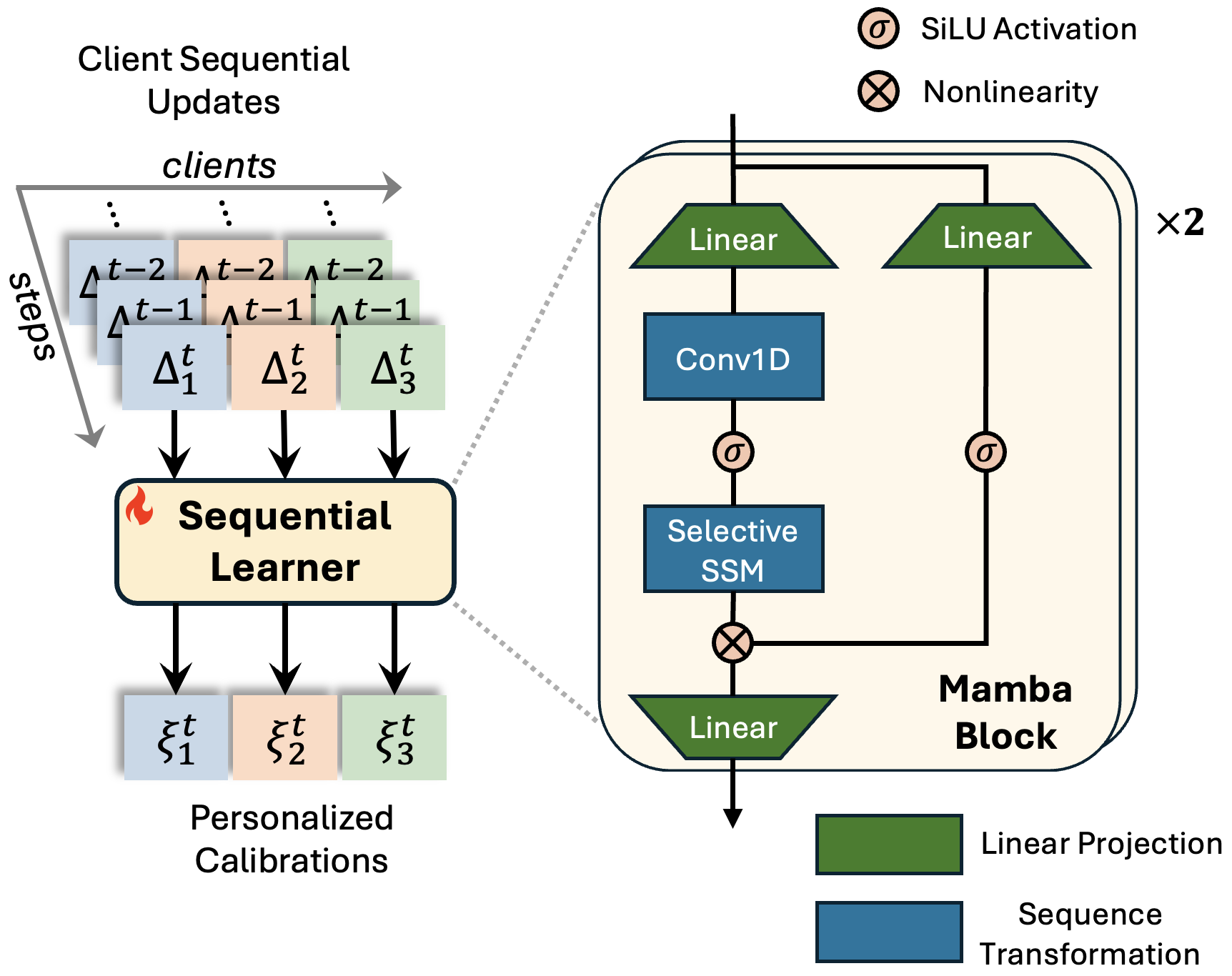} 
\caption{We implement our sequential learner using two Mamba blocks, where each Mamba block consists of linear projections, a 1D convolution, Selective SSM and a residual connection.}
\label{fig:full_archi}
\vspace{-2mm}
\end{figure}

In practice, we adopt two layers of Mamba as our sequential learner, as shown in Figure \ref{fig:full_archi}. Following the implementation by \citet{gu2023mamba}, we insert a normalization layer and a residual connection before and after each Mamba block. Within each Mamba block, the input is first linearly projected to a higher dimension and split into two parts, where the first part undergoes 1D depthwise convolution along the step dimension followed by a SiLU activation and Selective SSM, and the second part is for residual connection with the Selective SSM output.

\subsection{Complexity Analysis}
By adopting Mamba with Selective SSM as the core mechanism for sequence processing, we now analyze how its space complexity (i.e., model parameter size, intermediate activations during execution) depends on various input dimensions. Recall that we construct the input to the sequential learner as  $\boldsymbol{\Delta}^{t-L+1:t} \in \mathbb{R}^{D \times N \times L}$, where $D$ is the adapter's parameter size, $N$ is the number of clients, and $L$ is the sequence length. Since the adapter's parameter size is along the batch dimension, the size of the sequential learner is independent of the adapter's size. This is in contrast to other hypernetwork-based methods, which aim to learn the mapping in a high-dimensional parameter space, resulting in the hypernetwork's size increasing with the size of the client's network. In our case, the size of the learnable Mamba parameters depends only on the number of clients $N$, which scales far less significantly than the adapter's (or any target network's) dimension $D$. The internal operations along the step dimension (i.e., 1D convolution and Selective SSM) also result in memory consumption increasing linearly with the sequence length $L$. To avoid extensive memory usage of computing a large $D$ along the batch dimension, one can perform equivalent batch updates in practice, e.g., using a batch size of 32 and performing $D/32$ update steps.\

Regarding time complexity, although the selection mechanism prevents the parallel convolution of SSM, Selective SSM adopts an efficient parallel scan algorithm to replace the sequential recurrence, which effectively reduces the number of necessary sequential steps from $L$ to $2 \times \log_2(L)$, i.e., the time complexity is lowered to log-linear.

\section{B. Implementation Details}

\subsection{Data Split and Preprocessing}
\begin{figure}[h]
\centering
\includegraphics[width=\columnwidth]{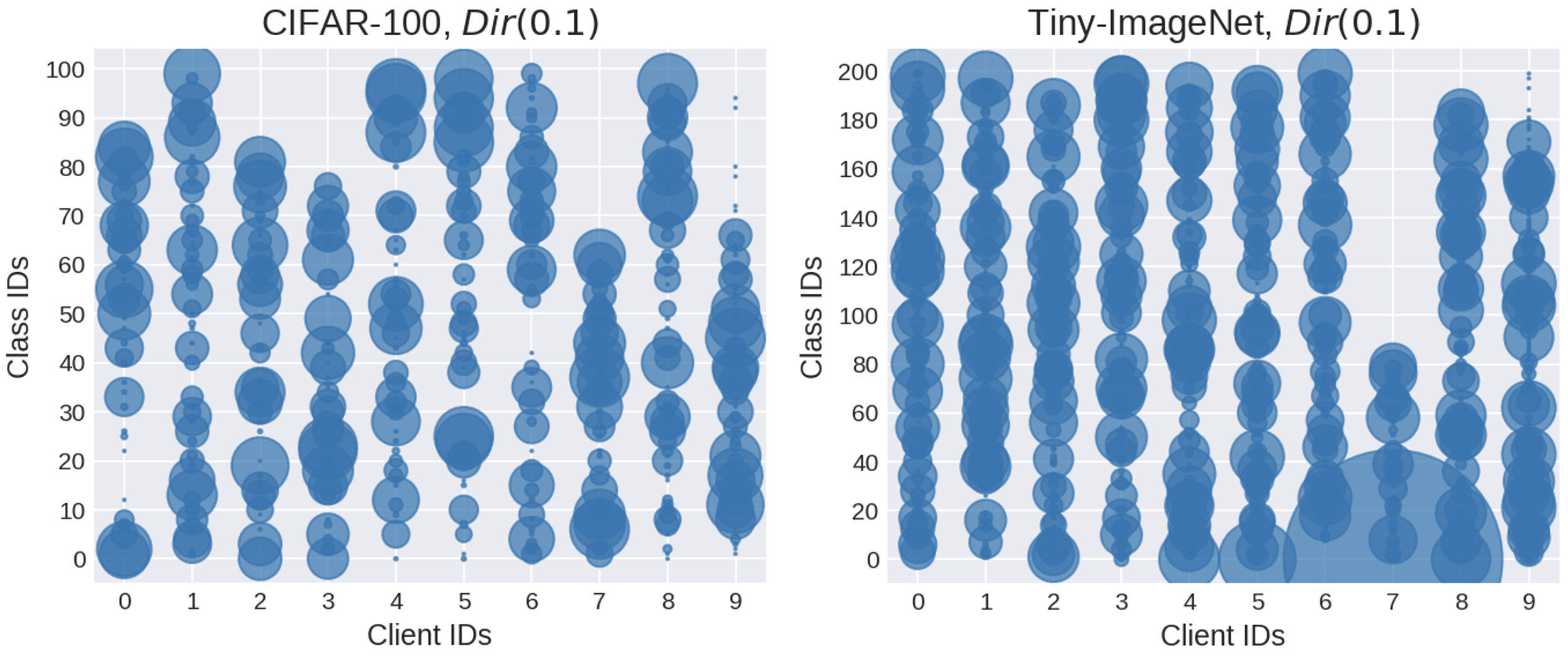} 
\caption{Visualization of data distributions among 10 clients for CIFAR-100 and Tiny-ImageNet, following a practical non-IID setting with $Dir(0.1)$.}
\label{fig:data_dist}
\end{figure}

We simulate three data heterogeneity scenarios using four datasets. CIFAR-100 and Tiny-ImageNet are used to simulate the label-skew scenario by distributing data from each class to $N$ clients following a Dirichlet distribution $Dir(\alpha)$ with $\alpha = 0.1$, which is commonly referred to as a practical non-IID setting \cite{zhangpersonalized,zhang2023fedala}. Figure \ref{fig:data_dist} shows the data distributions for CIFAR-100 and Tiny-ImageNet among 10 clients used in our experiments. DomainNet dataset is used to simulate the feature-skew scenario by treating each domain as a client, and Omniglot dataset is used to simulate a real-world Internet of Things where each individual is a client.

For CIFAR-100, Tiny-ImageNet, and Omniglot, we follow \citet{zhang2023fedala} to split the local data for each client into training and test sets with a 25/75 ratio. For DomainNet, we conduct experiments using the top ten most common classes and follow the train-test split from \citet{lifedbn}. We hold out 10\% of the training data as the validation set.\

All input images are resized to $224 \times 224$ with bicubic interpolation to match the input dimensions of ViT-B/16. Each image tensor is normalized with a mean of 0.5 and a standard deviation of 0.5 for all channels.

\subsection{Hyperparameter Settings}
The experimental pipeline for baseline comparisons was developed based on \citet{zhang2023fedala}'s PFLlib codebase. For all baselines, we tune the hyperparameters by performing grid search from a set of values on held-out validation set.

For Per-FedAvg and pFedMe, due to the additional updates for the inner-loop optimization, the local learning rate is a sensitive hyperparameter. We tune the local learning rate in \{0.001, 0.005, 0.01, 0.05\} for all four datasets, setting it to 0.01 for DomainNet and Omniglot, and to 0.005 for CIFAR-100 and Tiny-ImageNet.\

For APPLE, we set the learning rate for the aggregation weights (also termed the directed relationship vector) to 0.001, after tuning from \{0.001, 0.01, 0.1, 1.0\}. For FedALA, we apply adaptive local aggregation (ALA) to the adapters attached at all layers of the backbone. The learning rate for ALA is set to 0.1, tuned from \{0.001, 0.01, 0.1, 1.0\}, and the convergence threshold (i.e., the standard deviation of adjacent losses) is set to 1.0 to avoid long training time of ALA optimization.\

For FedRep, we follow the key idea of learning feature representations globally and the classification head locally by performing global aggregation on the adapters attached to all layers of the backbone (i.e., the feature extractor) and learning the classification heads locally for each client. For Ditto and PerAda, we set the coefficient for the proximal term to 0.1, tuning from \{0.01, 0.1, 1.0, 10.0\}. For PerAda, knowledge distillation (KD) is performed using CIFAR-10 \cite{krizhevsky2009learning} as the auxiliary dataset. The number of KD steps for training the global adapter is set to 100 for a better trade-off between computational cost and performance, and the learning rate for KD is set to 0.005, tuned from \{0.001, 0.005, 0.01, 0.05\}.

For pFedHN, pFedLA, and PeFLL, the hypernetworks for generating the adapter's parameters or the layer-wise aggregation weights are adopted from their original implementations. pFedHN employs a 4-layer MLP to compute the clients' embeddings from a set of learnable clients' descriptor vectors and a fully connected layer to generate the adapter's parameters at each backbone layer. The hidden size is set to 100. pFedLA adopts a similar hypernetwork architecture as pFedHN but assigns a unique hypernetwork to generate each client's embedding and uses the last fully connected layer to compute the layer-wise aggregation weights for combining client-specific adapters. For PeFLL, an embedding network is used to generate the clients' descriptor vectors based on the raw local data, which adopts a LeNet-style ConvNet with 2 convolutional layers (with filter size $5 \times 5$ and output channel size 32) followed by 3 fully connected layers. The hypernetwork for generating the adapter's parameters is similar to pFedHN. For all three hypernetwork-based baselines, the learning rate is tuned in \{0.0001, 0.0005, 0.001, 0.005, 0.01\}, and set to 0.001 for CIFAR-100, Tiny-ImageNet and DomainNet, and 0.0005 for Omniglot. Adam optimizer is used for updating the hypernetworks in all three methods.\

For our \texttt{pFedSeq}, we employ a 2-layer Mamba as our sequential learner. The expand factor of the linear projection within each Mamba block is set to 2, and the kernel size of the 1D convolution is set to 4. We set the expanded state dimension $M$ to 16, tuned from \{4, 8, 16, 32\}. The maximum sequence length $L$ is set to $20$ for CIFAR-100 and Tiny-ImageNet, and $10$ for DomainNet and Omniglot, tuned from \{5, 10, 15, 20, 25, 30\}, taking into account a better trade-off between performance and computational efficiency. We use Adam optimizer for training the sequential learner and tune the learning rate similarly to the other hypernetwork-based methods.

\label{appx:imp}

\section{C. Additional Results}
\label{appx:exp}

\subsection{Computational Efficiency of Baselines}
\begin{table}[h]
\setlength{\tabcolsep}{2pt}
\small
\centering
\begin{tabular}{l | c c c c}
\toprule
\multirow{2}{*}{\small\textbf{Method}} & \multicolumn{2}{c}{\scriptsize\textit{Label-Skew}} & \scriptsize\textit{Feature-Skew} & \scriptsize\textit{Real-World}\\
\cmidrule(lr){2-3} \cmidrule(lr){4-4} \cmidrule(lr){5-5}
&  \small CIFAR-100 & \small Tiny-ImageNet & \small DomainNet & \small Omniglot \\
\midrule
Local & 754.7 & 1720.0 & 197.7 & 479.7 \\
FedAvg & 758.8 & 1720.7 & 197.8 & 480.5\\
\midrule
\rowcolor{gray!20} \multicolumn{5}{l}{\scriptsize\textit{Meta-Learning-Based}} \\
Per-FedAvg & 763.7 & 2163.8 & 242.3 & 607.0\\
pFedMe & 3796.5 & 7040.8 & 989.8 & 2254.7\\
\midrule
\rowcolor{gray!20} \multicolumn{5}{l}{\scriptsize\textit{Personalized-Aggregation-Based}} \\
APPLE & 818.4 & 2167.3 & 269.1 & 656.5\\
FedALA & 1529.2 & 3096.7 & 474.3 & 892.7\\
\midrule
\rowcolor{gray!20} \multicolumn{5}{l}{\scriptsize\textit{Personalized-Network-Based}} \\
FedRep & 1512.1 & 3441.4 & 395.5 & 960.8\\
Ditto & 1425.5 & 3436.2 & 452.7 & 1048.2\\
PerAda & 1735.8 & 3753.8 & 655.3 & 1622.2\\
\midrule
\rowcolor{gray!20} \multicolumn{5}{l}{\scriptsize\textit{Hypernetwork-Based}} \\
pFedHN & 758.4 & 1721.0 & 198.3 & 476.4\\
pFedLA & 760.3 & 1722.6 & 204.8 & 482.1\\
PeFLL & 824.1 & 1805.3 & 270.1 & 531.3\\
\midrule
pFedSeq & 761.3 & 1721.7 & 198.3 & 483.6\\
\bottomrule
\end{tabular}
\caption{Average time taken (in seconds) per communication round, including local training time at client, as well as aggregation and any additional computations at server. All the experiments are performed using one NVIDIA A100 GPU.
}
\label{tbl:time}
\end{table}

In Table \ref{tbl:time}, we report the average time taken (in seconds) per communication round for each method on four datasets. The time accounts for local training at client, as well as aggregation and any additional computations at server.
Therefore, we use the average time taken per round to reflect and compare the computational efficiency.
From the results, we can see that our \texttt{pFedSeq} is among the most efficient PFL methods (comparable to pFedHN and pFedLA). Compared to the Local baseline, our \texttt{pFedSeq} achieves significant performance gains with only a few additional seconds. The extra time taken is mainly due to the update (i.e., backward pass) and inference (i.e., forward pass) on the sequential learner at the server, where both processes are performed once per communication round (i.e., a single-step full-batch update and a single forward pass). \

\begin{table*}[h]
\setlength{\tabcolsep}{6pt}
\scriptsize
\centering
\begin{tabular}{l | c  c  c  c}
\toprule
\multirowcell{2}{\textbf{Method}} & \multirowcell{2}{\textbf{Data Distribution}} & \multirowcell{2}{\textbf{Backbone}} & \multirowcell{2}{\textbf{Communicated} \\ \textbf{Parameter Size}} & \multirowcell{2}{\textbf{CIFAR-100} \\ \textbf{Performance}} \\
\\
\midrule
\rowcolor{gray!20} \multicolumn{5}{l}{\textit{Personalized-Aggregation-Based}} \\
FedFomo \cite{huang2021personalized} & 15 clients, pathological non-IID & ConvNet-based & $\sim$104k & 40.94\\
FedAMP \cite{zhangpersonalized} & 100 clients, practical non-IID & ResNet18 & $\sim$6.4m & 54.27\\
FedALA \cite{zhang2023fedala} & 20 clients, practical non-IID ($Dir(0.1)$) & ConvNet-based & $\sim$104k & 55.92\\
\midrule
\rowcolor{gray!20} \multicolumn{5}{l}{\textit{Personalized-Network-Based}} \\
FedRep \cite{collins2021exploiting} & 100 clients, pathological non-IID & ConvNet-based & $\sim$60.8k & 56.10\\
pFedLoRA \cite{yi2023fedlora} & 10 clients, pathological non-IID & Trainable ConvNet-based + LoRA Tuning & $\sim$35k & 75.58\\
\midrule
\rowcolor{gray!20} \multicolumn{5}{l}{\textit{Hypernetwork-Based}} \\
pFedHN \cite{shamsian2021personalized} & 10 clients, pathological non-IID & ConvNet-based & $\sim$104k & 68.15\\
L2C \cite{li2022learning} & 100 clients, pathological non-IID & ConvNet-based & $\sim$104k & 59.00\\
pFedLA \cite{ma2022layer} & 10 clients, pathological non-IID & ConvNet-based & $\sim$104k & 47.22\\
PeFLL \cite{scott2023pefll} & 100 clients, pathological non-IID & ConvNet-based & $\sim$128k & 56.00\\
pFedPG \cite{yang2023efficient} & 10 clients, practical non-IID ($Dir(0.1)$) & Frozen ViT-B/16 + Prompt Tuning & $\sim$7.68k & 55.91\\
\midrule
pFedSeq (Ours) & 10 clients, practical non-IID ($Dir(0.1)$) & Frozen ViT-B/16 + LoRA Tuning & $\sim$73k & 95.30\\
\bottomrule
\end{tabular}
\caption{Results of PFL baselines on the commonly used \textbf{CIFAR-100} dataset as reported in their original papers. We also record the setups used to obtain the results. Note that we only include PFL baselines here \underline{which have used CIFAR-100} in their experiments and have reported the results.}
\label{tbl:orig_settup}
\vspace{-3mm}
\end{table*}

As compared to other PFL methods, our \texttt{pFedSeq} is significantly more efficient than FedRep, Ditto and PerAda, which require additional training at local clients to separately learn the local and global components or to optimize the proximal term. PerAda performs additional training of the global model at server through distillation on an auxiliary dataset, making it the most costly among the three. Our \texttt{pFedSeq} is also more efficient than the learning-based aggregation methods APPLE and FedALA, which require explicit optimization of the personalized aggregation weights on clients' local data. pFedMe is the most costly among the compared PFL methods, due to the multi-step inner-loop optimization performed at each local update. Hypernetwork-based methods are arguably the most efficient, as the updates on the hypernetwork can be computed efficiently using feedback returned by clients, and the personalized adapters are generated conveniently by a single forward pass. \

As compared to the hypernetwork-based counterparts like pFedHN and pFedLA, our \texttt{pFedSeq} which operates on sequences of inputs is only slightly slower (note that the computation time of \texttt{pFedSeq} reported here is based on $L=10$). Thanks to the efficient parallel scan of Selective SSM to replace the sequential recurrence computations, the computational time of \texttt{pFedSeq} is reduced from linear in sequence length $L$ to log-linear, allowing it to handle long sequences with only a slight increase in computation time. Also, by directly using the stored model updates as inputs, \texttt{pFedSeq} is more efficient than PeFLL, which requires additional training of clients' descriptors based on clients' local data. Overall, our \texttt{pFedSeq} ranks 2nd or 3rd among eleven PFL methods in terms of computational efficiency, while achieving the best performance (with significant margins over the second-best performers pFedHN or PeFLL).


\subsection{Comparison of PFL Baselines in Original Setup}

In our experiments, we adapt all PFL baselines to the same federated adapter tuning setup (i.e., fine-tuning a frozen ViT with LoRA) to ensure fair comparisons. Here, we provide the results of PFL baselines on CIFAR-100 as reported in the literature under the original setups used by the authors. The setups differ in the backbone model used at local clients, and the way the data is distributed among clients. These PFL baselines also have different communication costs depending on the backbone model used and the requirements of the PFL algorithms.\

In Table \ref{tbl:orig_settup}, we record the data distribution method, the backbone used, the communicated parameter size, and the performance on CIFAR-100 as reported in their original papers for several PFL baselines. Generally, the local test performance is largely determined by the capability of the backbone model, where our setup of tuning LoRA on a powerful, pre-trained ViT results in much better performance as compared to most of the PFL baselines that utilize ConvNet-based model as the backbone. Unlike the traditional setup of updating the full backbone model, federated adapter tuning, which updates only the lightweight adapter, leads to a much smaller parameter size to be transmitted and reduced computational overhead at clients. Perhaps the most similar setup to ours is the recent work pFedPG, which distributes data among 10 clients using $Dir(0.1)$ and leverages a pre-trained ViT-B/16. Note that pFedPG fine-tunes the ViT using prompt tuning while ours employs LoRA adapter tuning. By learning only the personalized prompts for each client, pFedPG maintains the smallest communicated parameter size. However, this approach also limits the effectiveness of adapting the ViT to local clients, resulting in significantly poorer performance compared to adapter tuning. Overall, Table \ref{tbl:orig_settup} showcases the advantages of adopting adapter tuning in PFL, which achieves a better trade-off between communication costs (as well as local computation costs) and local personalization performance.

\subsection{Additional Figures}
This section includes learning curve plots (of baselines or variants) for the various studies discussed in the main paper.\

\begin{figure}[h]
\centering
\includegraphics[width=\columnwidth]{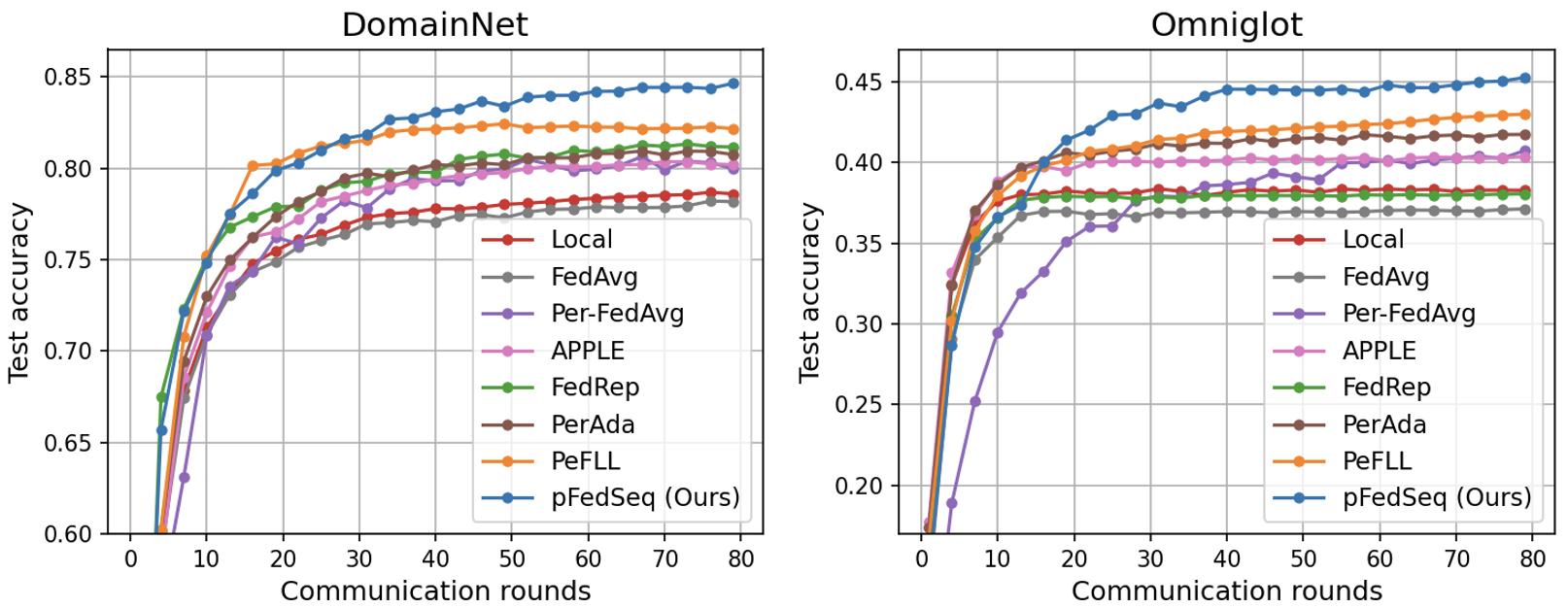} 
\caption{Learning curves of representative PFL baselines on DomainNet and Omniglot.}
\label{fig:overall_perf_1}
\end{figure}

Figure \ref{fig:overall_perf_1} shows the learning curves of \texttt{pFedSeq} and representative PFL baselines on DomainNet and Omniglot. Similar to Figure \ref{fig:overall_perf_2}, \texttt{pFedSeq} exhibits comparable performance to FedRep during the warm-up phase (i.e., the first 10 rounds) and quickly surpasses all the baselines within 15 rounds on DomainNet and 5 rounds on Omniglot, demonstrating the effectiveness of the personalized adapters generated by \texttt{pFedSeq} in facilitating faster learning and boosting performance.

\begin{figure}[h]
\centering
\includegraphics[width=\columnwidth]{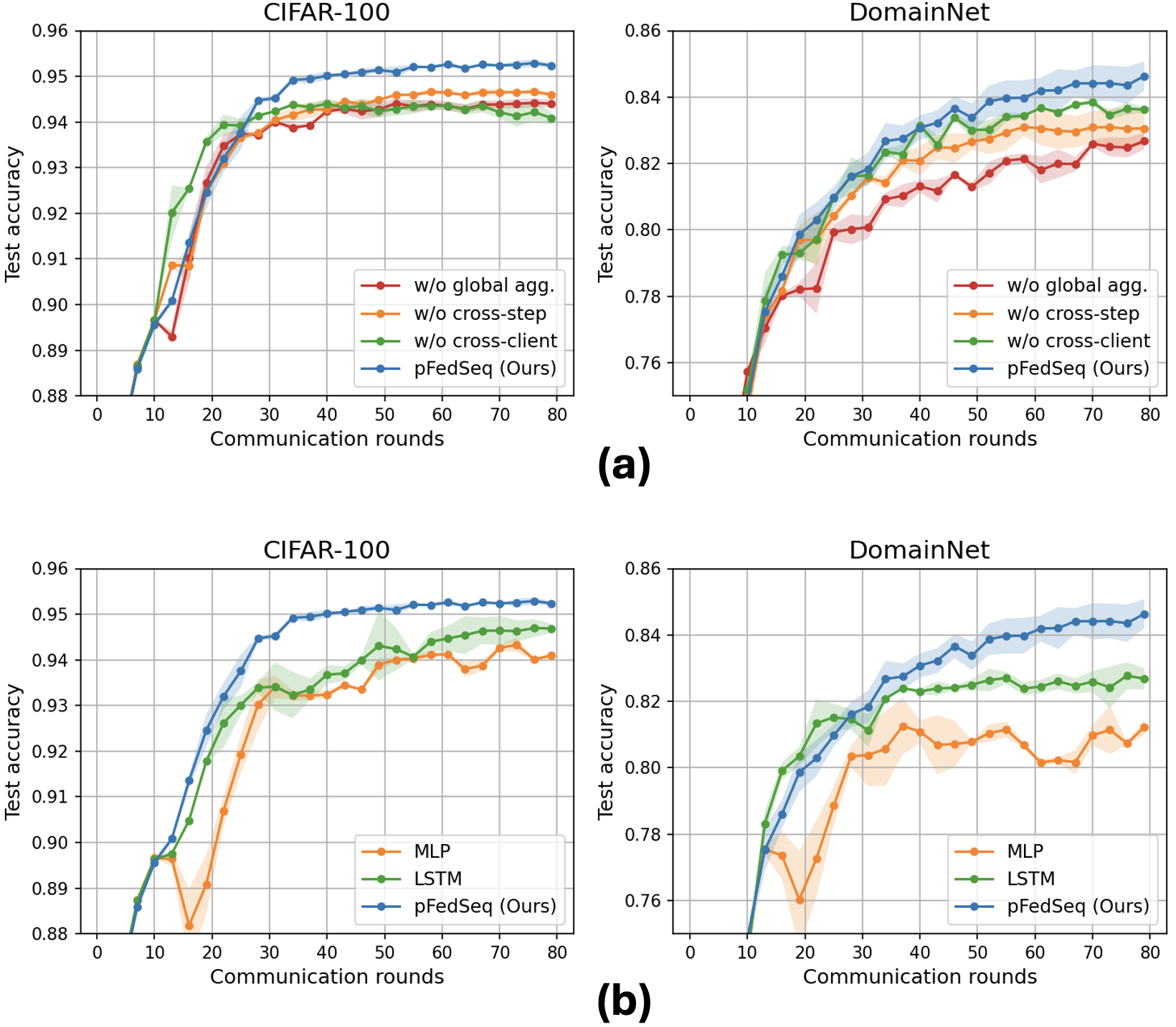} 
\caption{Learning curves for analyzing the effectiveness of (a) the key components and (b) the learner architecture.}
\label{fig:learner_archi_curves}
\end{figure}

Figure \ref{fig:learner_archi_curves}a presents the learning curves of the three variants introduced in the analysis of key components on CIFAR-100 and DomainNet (see Table~\ref{tbl:key_component} in the main paper for a recall of these three variants). From the plots, we observe that for DomainNet, removing global aggregation (red line) significantly degrades performance and leads to unstable learning. On CIFAR-100, the abrupt drop of the red line after the warm-up phase (when switching from the global adapter to personalized adapters directly generated by the sequential learner) also demonstrates the importance of global aggregation in stabilizing the learning process. Removing cross-client modeling (green line) and focusing on clients' own update trajectories may lead to faster learning in the earlier phase after warm-up. However, this variant is eventually outperformed by \texttt{pFedSeq}, indicating the importance of considering cross-client knowledge transfer to alleviate overfitting to local data and promote better generalization. Lastly, removing cross-step modeling (orange line), equivalent to setting $L=1$, results in less stable learning and poorer performance at convergence compared to our \texttt{pFedSeq}, which considers a longer sequence of updates for generating personalized adapters (here, we set $L=20$ for CIFAR-100 and $L=10$ for DomainNet). This demonstrates the effectiveness of incorporating past updates for improved learning.\

Figure \ref{fig:learner_archi_curves}b shows the learning curves of two variants using alternative architectures: MLP and LSTM, for the sequential learner (see Table~\ref{tbl:learner_archi} in the main paper for a recall of these two variants). From the plots, we can see that employing an MLP-based network (orange line) seriously degrades the learning stability. This is mainly due to that, by simply concatenating the sequence inputs along a single dimension, the temporal information inherent in the inputs at various steps is lost, leading to less effective modeling of the sequential dependencies. Moreover, treating a sequence of inputs as one long token and processing it with fully connected layers is less efficient, giving rise to more complex modeling and adversely affecting learning stability. Utilizing an LSTM-based network (orange line) for the sequential learner leads to more stable and improved sequence modeling performance compared to the MLP-based network. However, reusing the same cell across all steps limits the model's capacity to discern valuable information from the context, whereas Selective SSM with its input-dependent selection mechanism effectively addresses this issue, leading to enhanced performance. Moreover, by replacing sequential recurrence with parallel scan, Selective SSM is also more efficient than traditional RNN-based networks like LSTM which suffer from linear time complexity.\




\begin{figure}[h]
\centering
\includegraphics[width=0.6\columnwidth]{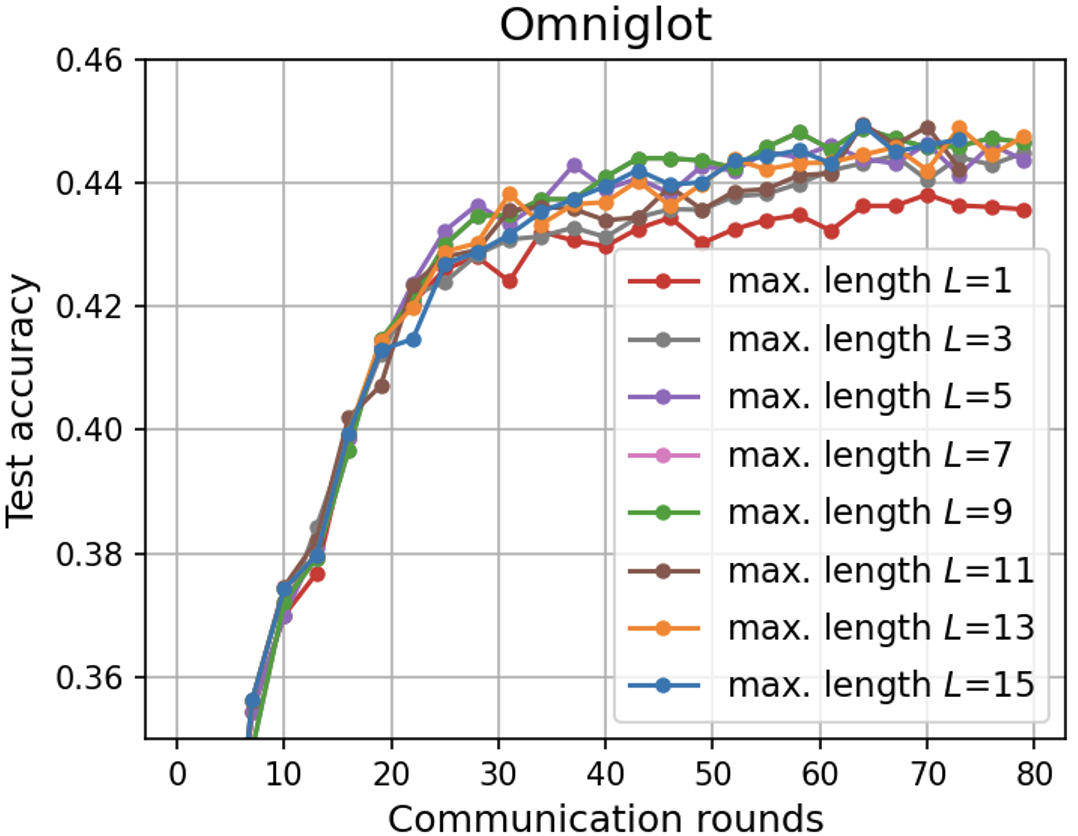} 
\caption{Learning curves for different values of maximum sequence length $L$ on Omniglot.}
\label{fig:seq_curves}
\end{figure}

\begin{figure}[h]
\centering
\includegraphics[width=0.9\columnwidth]{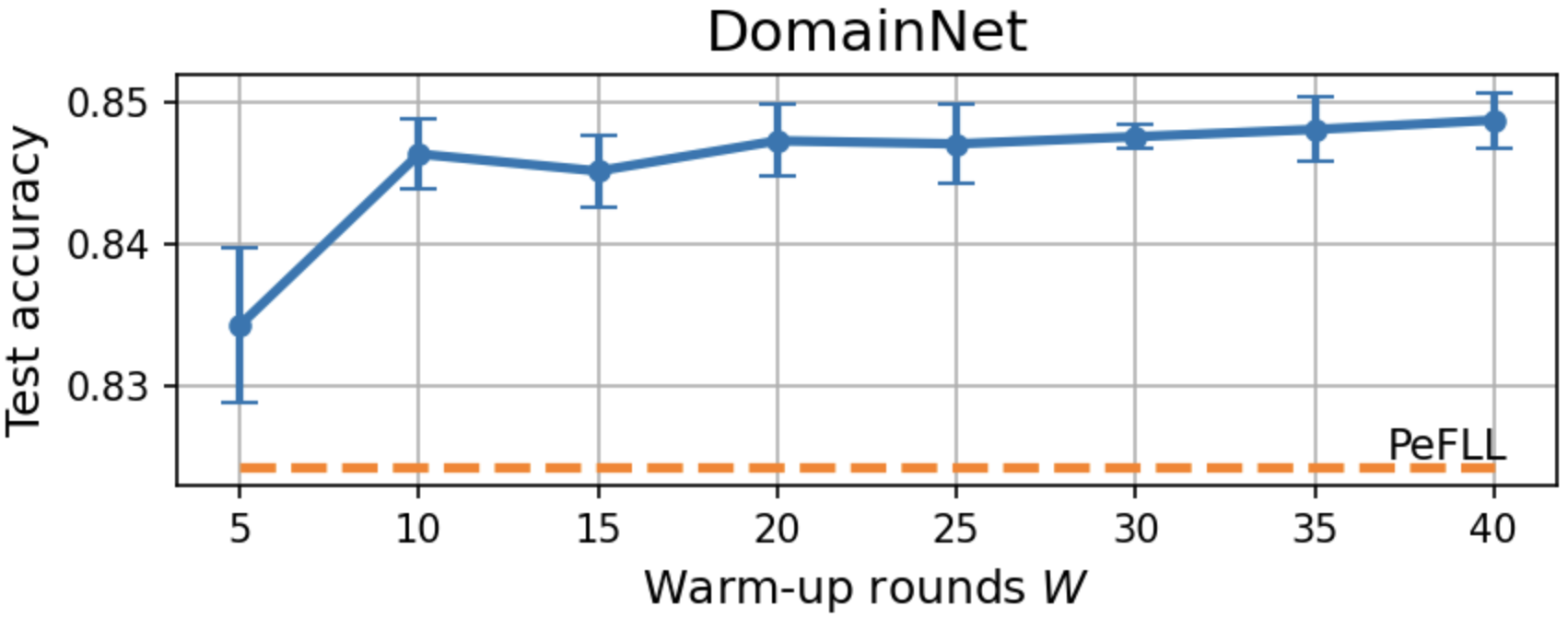} 
\caption{Performance of \texttt{pFedSeq} by varying the number of warm-up rounds $W$ on DomainNet.}
\label{fig:warm_up_plot}
\end{figure}

\begin{figure}[h]
\centering
\includegraphics[width=0.6\columnwidth]{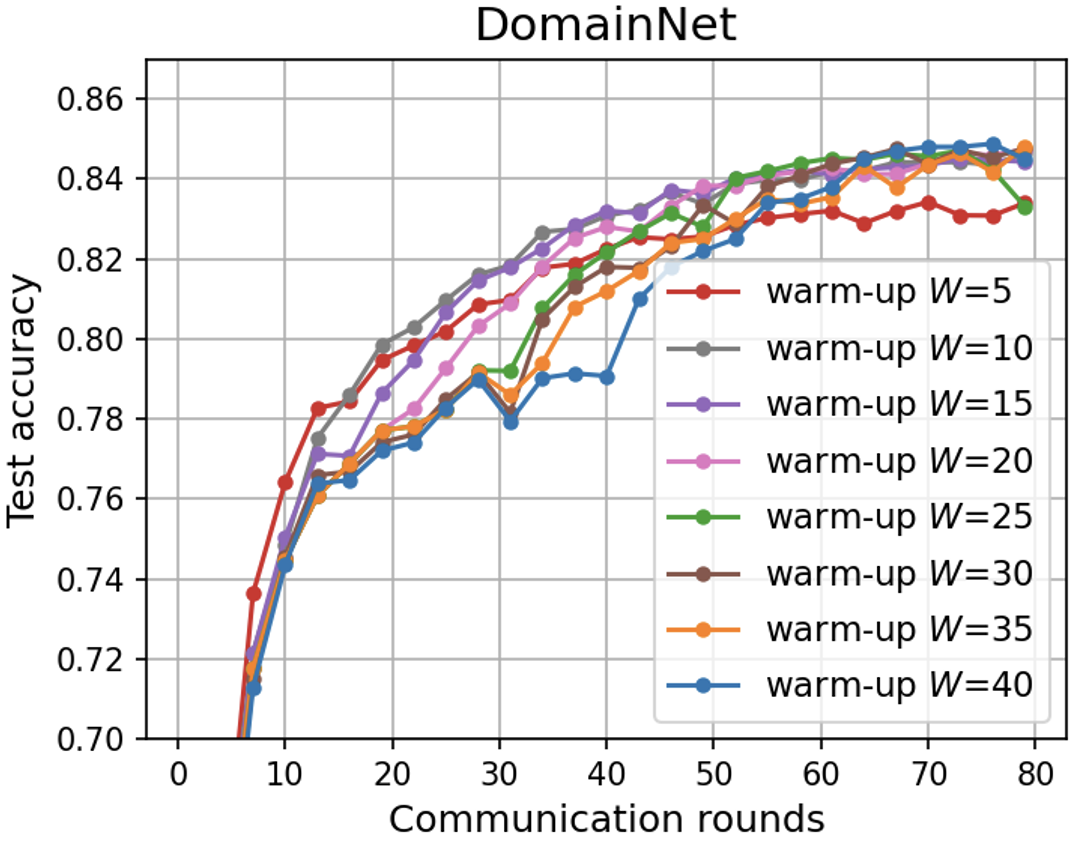} 
\caption{Learning curves for different values of warm-up rounds $W$ on DomainNet.}
\label{fig:warm_up_curves}
\end{figure}

In our framework, we allocate a certain warm-up period $W$ to allow the sequential learner to be sufficiently trained before applying it to generate personalized adapters for the next round. We investigate the impact of $W$ by varying it in \{5, 10, 15, 20, 25, 30, 35, 40\} and testing the performance on DomainNet. The results are shown in Figure \ref{fig:warm_up_plot}. First, we can see that without sufficiently warming up the sequential learner (e.g., $W=5$), the performance attained is relatively lower. This is because utilizing an inadequately trained sequential learner to generate personalized adapters may adversely affect the next round of local updates and the subsequent global aggregation, leading to suboptimal convergence. Also, applying a short warm-up period can lead to less stable training, as evidenced by the larger variance.
Instead, increasing warm-up period to 10 leads to 1.21\% increase in performance and 0.30\% drop in standard deviation compared to $W=5$. Though further increasing the warm-up period slightly improves the final performance (i.e., performance attained at the 80-th communication round), 
we set the warm-up period $W$ to 10 in all our experiments to avoid prolonged warm-up, during which clients can only receive the less capable global adapter for local evaluation and deployment. Also, a long warm-up period is not favorable for convergence efficiency. 
Overall, our \texttt{pFedSeq} outperforms PeFLL for all the values of $W$ tested. A plot of the learning curves for different values of $L$ is shown in Figure \ref{fig:warm_up_curves}.

\end{document}